\documentclass[runningheads]{llncs}

\usepackage{eccv}

\usepackage{eccvabbrv}
\usepackage{subcaption}
\usepackage{makecell}
\usepackage{graphicx}
\usepackage{booktabs}

\usepackage{multirow}
\usepackage{mathtools}
\usepackage[accsupp]{axessibility}  % Improves PDF readability for those with disabilities.
\usepackage{tikz}
\usetikzlibrary{arrows.meta}
\usepackage{cancel}
\usepackage{array}
\usepackage[table]{xcolor}
\usepackage{pifont}
\newcommand{\cmark}{\ding{51}}
\newcommand{\xmark}{\ding{55}}
\usepackage{hyperref}

\usepackage{orcidlink}

\newcommand{\eunyi}[1]{{\color{cyan}{#1}}}

\begin{document}

% \title{Does Visual Expressiveness Help Domain Generalization? A Text-anchored Bottleneck} 
% \title{Text-Anchored Information Bottleneck for Domain Generalization} 
\title{Domain Generalization via Text-Anchored Information Bottleneck} 


% Second Author\inst{2,3}\orcidlink{1111-2222-3333-4444} \and
% Third Author\inst{3}\orcidlink{2222--3333-4444-5555}}
\author{Eunyi Lyou \and
Yunjeong Choi \and
Junho Lee \and
Joonseok Lee\thanks{Corresponding author}}

\authorrunning{E.~Lyou et al.}

\institute{Seoul National University, Seoul, Republic of Korea \\
\email{\{onlyou0416, racheal0, joon2003, joonseok\}@snu.ac.kr}}


\maketitle

\begin{abstract}
Visual recognition models often fail when deployed in new environments. Domain Generalization (DG) addresses this by learning representations that remain invariant to environment-specific variations. Recent approaches increasingly rely on large vision-language models, assuming that preserving their expressive visual representations improves robustness. However, we show that such visual expressiveness can instead propagate spurious cues that tie representations to the training environments, hindering invariant learning. We therefore discard visual guidance and instead treat the language embedding space as the primary source of domain invariance, naturally acting as an information bottleneck that preserves core semantics while suppressing domain-specific variations. Extensive experiments across diverse backbones exhibit state-of-the-art performance and further analyze what makes guidance effective for robust generalization. These findings shift the focus of DG from improving representations to designing supervision that enforces invariance.
\end{abstract}

\section{Introduction}
\label{sec:1-intro}
% https://docs.google.com/presentation/d/1R6CGa3xkgWLrSoNvPaLKou8v4iTYgqGhoft0sMkxiH4/edit?slide=id.g3bd849e84a3_0_4#slide=id.g3bd849e84a3_0_4
\begin{figure}[tb]
  \centering
  \includegraphics[width=\linewidth]{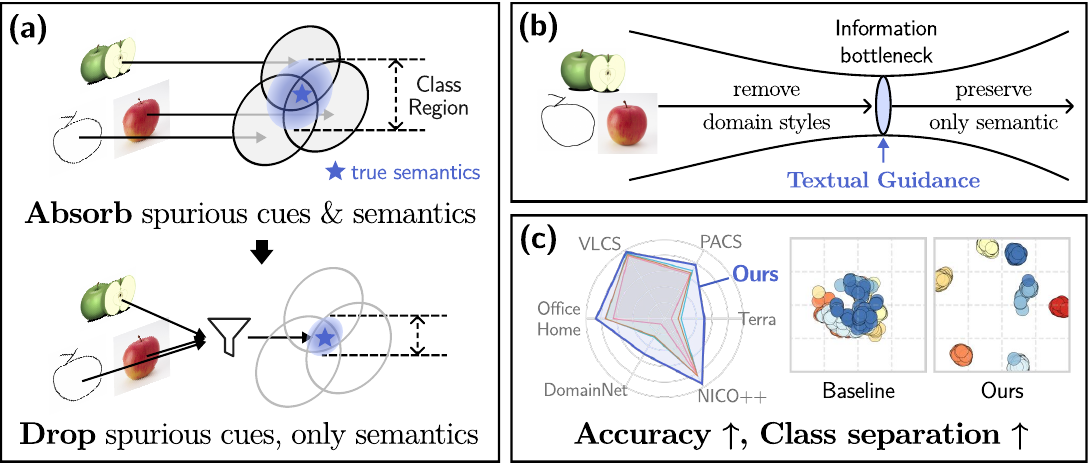}
  \caption{(a) Visual encoders inevitably absorb spurious domain cues alongside domain-invariant semantics. This inflates class regions and blurs boundaries, hindering robustness. Ideally, models should drop domain-specific variations while preserving only core semantics. (b) Redefining invariance via a text-anchored Information Bottleneck (IB). Textual guidance acts as a semantic filter, preserving information shared between text and image as core semantics while dropping non-shared domain styles. (c) Our method achieves state-of-the-art performance across DG benchmarks, producing an embedding space with improved class separation.}
  \label{fig:summ}
\end{figure}

% Task
Despite progress in computer vision, visual recognition models often fail to maintain performance under environmental changes at deployment, when test samples deviate from training distribution~\cite{NIPS2006_b1b0432c, 10.5555/1462129}.
%Such failures arise when the \textit{i.i.d.} assumption—rarely satisfied in real-world applications—is violated~\cite{NIPS2006_b1b0432c, 10.5555/1462129}.
Unlike domain adaptation~\cite{NIPS2006_b1b0432c, 36364}, which has access to target distribution during training,
%While domain adaptation~\cite{NIPS2006_b1b0432c, 36364} assumes access to data from the deployment setting during training,
domain generalization (DG)~\cite{pmlr-v28-muandet13} addresses the stricter scenario of generalizing to unseen environments without any prior exposure.
Under this constraint, DG aims to learn domain-invariant representations that preserve only essential semantics of the input.

% Related Work
To pursue such invariance, standard DG frameworks train across multiple source domains, under the assumption that reducing distributional discrepancies among diverse sources will expose an underlying invariant subspace.
For instance, aligning a simple sketch with a real photograph of an apple is expected to reveal the shared semantic ``apple-ness'' beneath domain-specific variations.
Accordingly, traditional DG methods primarily align feature distributions across domains~\cite{pmlr-v28-muandet13, 10.1007/978-3-319-49409-8_35, pmlr-v37-ganin15, arjovsky2020invariantriskminimization, Ahuja2021InvariancePM, 10.1007/978-3-030-58607-2_12, Li2021InvariantIB, Yu2023INSUREAI, guo2023domaindrop} 
or employ robust training strategies to mitigate domain-specific correlations~\cite{10.1007/978-3-031-20050-2_26, Wang2023SharpnessAwareGM, Cha2021SWADDG}.

Recently, the field has turned to large-scale vision–language models (VLMs) such as CLIP~\cite{pmlr-v139-radford21a}, motivated by their strong zero-shot generalization ability.
Because standard fine-tuning can compromise this robustness under distribution shift~\cite{Pham2021CombinedSF}, current strategies aim to preserve the original zero-shot representations while adapting them to the DG task~\cite{Wortsman_2022_CVPR}—often freezing the text encoder while adapting the visual branch—via knowledge distillation~\cite{Huang2023ASS, Addepalli_2024_CVPR, CLIP-DTP}, prompt tuning~\cite{Zhou2022ConditionalPL, Zhou_2022, khattak2022MaPLe, Zhang2021DomainPL, 10.1007/s11263-023-01951-2, Cheng2024DisentangledPR, Liu2023TDGTD, CLIP-DTP}, or weight ensembling~\cite{Lew2023GradientEF, pmlr-v202-shu23a, Wortsman_2022_CVPR}.

% 같은 class에 속하는 object 끼리도 차이가 있게 마련인데, it can be either intra-class variation or domain gap.
  % 전자는 여러 domain에 걸쳐 공통적으로 나타나는 variation으로, 이건 그 class의 본질을 벗어나지는 않는 범위로 봐야 함.
  % 후자는 특정 domain에서만 나타나는 variation으로, class의 본질은 벗어난 것으로 봐야 하고, visual encoder가 이런 variation마저 해당 class의 feature로 배워버리면 DG에 방해가 되는 spurious feature임.
% 그런데 visual encoder를 학습할 때는 (명시적으로 여러 domain의 데이터를 대조해서 학습하지 않는 이상) variation 자체만 보고 이게 어느 쪽인지 알 수 없다.
% feature로 배울지 말지는 model의 capacity와 그러한 variation이 나타나는 빈도에 따라서 결정될 뿐, 무엇이 domain-specific하고 무엇이 일반적인 intra-class variation인지 구별하긴 어려움.
% 그래서, 명시적으로 multi-domain-aware하게 학습하지 않은 일반적인 visual encoder가 학습해 놓은 정보를 무조건적으로 보존하려고 하면 spurious feature도 흘러들어올 수밖에 없다.
% 그래서 우린 information bottleneck을 이용해서 이걸 컨트롤하려고 하고, 가장 확실하게 domain-invariant한 text signal만을 이용하자고 주장하는 것.
These approaches share a common premise that preserving the pre-trained visual knowledge of VLMs (\emph{e.g.}, CLIP) benefits generalization through zero-shot robustness or semantic expressiveness.
\textit{But does preserving the visual features truly benefit domain generalization?}
Consider the `apple' examples in~\cref{fig:summ}(a).
%\eunyi{While a highly expressive visual encoder successfully captures the core semantics of the object ($\bigstar$)},
Highly expressive visual encoders entangle the core semantics ($\bigstar$)
%it inevitably entangles them 
with visual styles, such as sketch strokes, photographic textures, or artistic abstractions.
Without an explicit criterion to distinguish true semantics from these variations, 
models preserving such visual representations tend to retain any feature that is predictive of labels in the training domains.
Consequently, domain-specific cues are absorbed alongside true semantics,
% the model inadvertently retains any feature that is \eunyi{predictive of the label in the training domains.}
%frequently observed from and predictive within 
% the training distribution.
% \joonseok{Consequently, a model guided by visual representations inevitably absorbs domain-specific signals, without learning to filter out those spurious cues.
expanding the range of a class beyond its core and blurring its boundary under domain shifts.
In contrast, the ideal representation retains only core semantics by dropping domain spurious cues.
To avoid inheriting domain entanglement from visual guidance, we propose to discard it altogether.
Instead, we elevate the text space to serve as the primary source of domain invariance, drawing on both the stability of its anchors and their semantic structure (\cref{sec:3-motivation}). 
Concretely, as shown in~\cref{fig:summ}(b), features from diverse domains are constrained to align with text-defined anchors through a class-conditional Information Bottleneck (IB). 
These anchors remove domain-induced styles not shared across modalities while preserving information shared between input images and text.
% This alignment is enforced through a class-conditional Information Bottleneck grounded in fixed text embeddings, which serve as semantic priors that preserve input-text shared information while suppressing domain-induced styles.
% \joonseok{To address this issue, we need} a mechanism to drop the domain-specific cues and preserve only the common semantics, as illustrated at the bottom of \cref{fig:summ}(a).
% Unlike conventional DG methods that infer invariance from correlations among training domains, \joonseok{we propose to control the training process by enforcing information bottleneck, as in \cref{fig:summ}(b).
% Specifically, we propose a DG framework that entirely drops highly expressive visual features as guidance.
% Instead, we adopt text embeddings as a stable semantic reference, \joonseok{which is almost certainly invariant} across domains.

% Results - fig. 1 (c)
Extensive experiments demonstrate that our approach consistently achieves the state-of-the-art performance across representative DG benchmarks with clearer separation in the learned embedding space, as illustrated in \cref{fig:summ}(c).
Importantly, our evaluation spans diverse backbone architectures, further demonstrating the generality of the proposed framework.

% Method - fig.1 (b)
%To address this issue, we propose a DG framework that entirely drops highly expressive visual features as guidance. Instead, we treat fixed text embeddings as a stable semantic reference across domains. Unlike conventional DG methods that infer invariance from correlations among training domains, our approach defines invariance through this fixed cross-modal anchor.
%As illustrated in~\cref{fig:summ}(b), we implement this idea through an Information Bottleneck (IB), where textual guidance serves as a semantic sieve to suppress domain-specific variations and preserve only information aligned with shared image–text semantics, yielding principled and tractable objectives for cross-modal purification.

Our contributions are summarized as follows:
\begin{enumerate}
  \item Revisiting the visual guidance in DG, we reveal that highly expressive visual encoders can propagate domain-specific cues and hinder domain invariance under distribution shift.
  %\item{\textbf{Revisiting visual expressiveness in DG.} We show that highly expressive visual encoders can propagate spurious cues and hinder domain invariance under distribution shift.}

  \item We propose a purely text-guided approach based on IB theory, suppressing domain-specific variations while preserving shared semantics.
  %\item{\textbf{Text-anchored Information Bottleneck for semantic purification.} We redefine invariance through a fixed cross-modal semantic anchor and realize it via a text-guided Information Bottleneck that suppresses domain-specific variations while preserving shared semantics.}
  
  \item{%\textbf{Consistent and architecture-agnostic generalization.}
  Through extensive experiments across diverse DG benchmarks and backbone architectures, we demonstrate consistent state-of-the-art performance and improved reliability under domain shift.}
  
  \item We further analyze guidance signals in DG and highlight supervision design as a key factor for learning invariant representations.
\end{enumerate}
\section{Related Work}
\label{sec:2-rel}
% https://docs.google.com/presentation/d/1R6CGa3xkgWLrSoNvPaLKou8v4iTYgqGhoft0sMkxiH4/edit?slide=id.g3bd849e84a3_0_4#slide=id.g3bd849e84a3_0_4

% \begin{figure}[t]
%   \centering
%   \includegraphics[width=\linewidth]{figures/benchmark.pdf}
%   \caption{\eunyi{Distribution shifts in standard DG benchmarks. To reflect real-world deployment, these datasets cover stylistic and conceptual shifts (\emph{e.g.}, art vs. photo), data acquisition shifts (\emph{e.g.}, camera location), and covariate background shifts.}}
%   \label{fig:benchmark}
% \end{figure}

% \subsubsection{Domain Generalization (DG).}
\textbf{Domain Generalization (DG).}
Visual recognition models often suffer from real-world distribution shift.
To evaluate these failures, established benchmarks span diverse shifts: style variation (\emph{e.g.}, sketch \textit{vs}. photo)~\cite{dataset-pacs, dataset-oh, dataset-dn}, differences in dataset acquisition process~\cite{dataset-vlcs} or camera location~\cite{dataset-ti}, and background shifts~\cite{dataset-nico}.
Together, these benchmarks expose the inherent brittleness of models under unseen environments.

% Visual recognition models often significantly suffer from distribution shift.
% To evaluate these failures, prior work (\cref{fig:benchmark}) has developed benchmarks spanning style variation~\cite{dataset-pacs, dataset-oh, dataset-dn}, data acquisition process differences~\cite{dataset-vlcs, dataset-ti}, and background shift~\cite{dataset-nico}, exposing brittleness of models under unseen environments.

DG seeks representations that remain invariant across such domain discrepancies without access to target data. 
%\subsubsection{Classical DG.}
Classical approaches primarily extract invariance by aligning source-domain feature distributions via statistical matching~\cite{8578664, 10.1007/978-3-319-49409-8_35}, 
adversarial learning~\cite{10.5555/2946645.2946704, pmlr-v37-ganin15, 8578664}, 
or feature disentanglement~\cite{10.1007/978-3-030-58545-7_18, Lv_2022_CVPR, Piratla2020EfficientDG, pmlr-v28-muandet13, Yu2023INSUREAI, guo2023domaindrop,jeon2022conservative,jeon2023unified}.
Others instead regularize optimization dynamics~\cite{arjovsky2020invariantriskminimization, Ahuja2021InvariancePM}, 
gradient constraints~\cite{Wang2023SharpnessAwareGM, Lew2023GradientEF}, 
ensembling~\cite{Cha2021SWADDG, Jain2023DARTDT, arpit2022ensemble, li2022domaingeneralizationusingpretrained}, 
and consistency guidance~\cite{10.1007/978-3-031-20050-2_26}.
These approaches define invariance based on patterns shared across the training domains, yet such shared structure often mix true semantics with spurious correlations, limiting generalization beyond source-like distributions.
% In contrast, we define invariance through a fixed, training-independent cross-modal semantic anchor.

%\subsubsection{CLIP-based DG.}
Recent approaches increasingly build upon CLIP~\cite{pmlr-v139-radford21a}, adapting its visual encoder while attempting to preserve zero-shot robustness.
Representative strategies include prompt optimization~\cite{Zhou2022ConditionalPL, Zhou_2022, khattak2022MaPLe, Zhang2021DomainPL, 10.1007/s11263-023-01951-2, Cheng2024DisentangledPR, Liu2023TDGTD, CLIP-DTP}, robust fine-tuning and weight ensembling~\cite{Lew2023GradientEF, pmlr-v202-shu23a, Wortsman_2022_CVPR, Nam2024LipsumFTRF}, and knowledge distillation~\cite{Huang2023ASS, Addepalli_2024_CVPR, CLIP-DTP}. 
Across these methods, the visual encoder is typically the component being updated during adaptation.
Meanwhile, its pretrained knowledge is preserved and reused through mechanisms such as regularization, ensembling, or distillation.

In contrast, the text encoder is usually kept frozen and serves as a relatively stable semantic reference. 
Depending on the method, text embeddings are used 
(i) as auxiliary alignment targets alongside visual supervision~\cite{Huang2023ASS, Addepalli_2024_CVPR, CLIP-DTP, pmlr-v202-shu23a}, 
(ii) as regularizers to constrain visual drift from the original zero-shot space~\cite{Nam2024LipsumFTRF}, or 
(iii) as diversified prompts to steer visual representations via multimodal alignment~\cite{Cheng2024DisentangledPR, Liu2023TDGTD, 10.1007/s11263-023-01951-2}.
We instead revisit the functional roles of CLIP’s visual and text encoders in DG, minimizing reliance on visual guidance and treating textual semantics as the primary basis for defining invariance.

\vspace{0.1cm} 
\noindent
\textbf{Information Bottleneck in DG.}
% \subsubsection{Information Bottleneck in DG.}
Empirical Risk Minimization (ERM)~\cite{Vapnik1998} often struggles under distribution shift due to spurious correlations. While Invariant Risk Minimization~\cite{arjovsky2020invariantriskminimization} enforces a shared classifier across environments, it does not fully eliminate such biases. To further constrain representations, several classical DG methods adopt the IB principle.
As the IB objective is intractable, these approaches rely on variational formulations, introducing KL-based regularization toward simple, typically non-semantic priors~\cite{Ahuja2021InvariancePM, Li2021InvariantIB}, with meta-learning \cite{10.1007/978-3-030-58607-2_12}, or coupling it with feature disentanglement objectives~\cite{Yu2023INSUREAI}.
In contrast, we anchor the bottleneck to fixed text embeddings as an explicit semantic prior.
% , so that compression is guided by stable label semantics rather than imposed through a task-agnostic prior.
Existing IB priors are either uninformative (\emph{e.g.}, $\mathcal{N}(0, \mathbf{I})$), giving no signal to separate semantic from spurious factors, or learned from source images, inheriting domain bias.
In contrast, our approach is class-conditional yet label-derived, signaling what to preserve without injecting domain bias.
\section{Motivation}
\label{sec:3-motivation}
% https://docs.google.com/presentation/d/1R6CGa3xkgWLrSoNvPaLKou8v4iTYgqGhoft0sMkxiH4/edit?slide=id.g3bd849e84a3_0_4#slide=id.g3bd849e84a3_0_4

To assess whether visual and textual modalities provide structurally reliable basis for domain-invariant guidance, we investigate the following questions:
1) Do their embedding spaces remain stable across domains? 2) How much domain-invariant and domain-specific information do the modality-specific encoders encode? 
3) How do visual and textual guidance signals affect learning dynamics?
% 3) How much does it actually impact on learning dynamics when we guide with domain-specific signals?

%\joonseok{In order to} identify \joonseok{if visual and textual modalities}
%which modality 
%provide structurally reliable basis for defining invariance, we first empirically examine how visual and textual representations behave under domain shift in~\cref{sec:3-1,sec:3-2}, and analyze their implications for domain generalization~\cref{sec:3-3,sec:3-4}.

\begin{figure}[t]
  \centering
  \includegraphics[width=\linewidth]{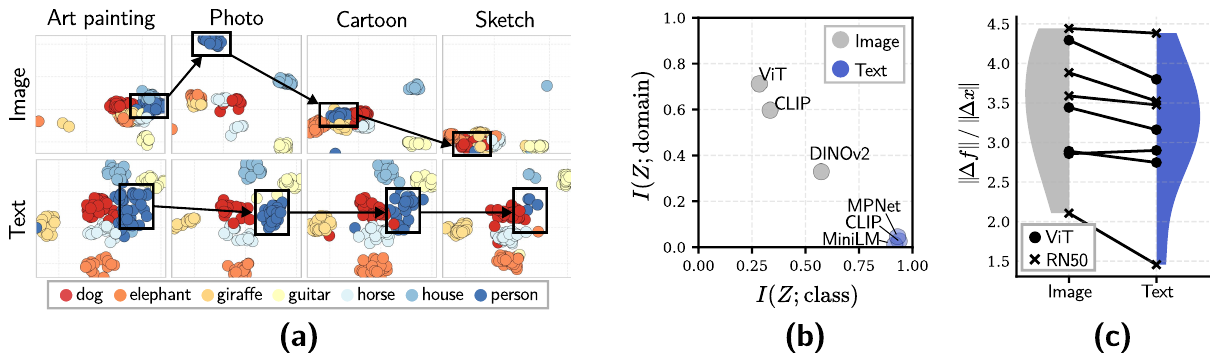}
  \caption{\textbf{Motivational experiments}. (a) Textual embedding space (bottom) shows superior stability across domains than visual counterpart (top). (b) Visual encoders (gray) tend to possess both domain-specific and core-class information, while textual ones (blue) contain only the latter. (c) Text-guided models tend to yield lower Lipschitz constants, indicating smoother and more stable representations.
  %\eunyi{Local Lipschitz constants under different guidances: visual yields sharper, less stable representations, while textual induces smoother ones.}
  }
  \label{fig:motiv}
\end{figure}

\subsection{Empirical Examination}
\label{sec:motivation:exp}

%\subsection{Embedding Geometry under Domain Shift}
%\label{sec:3-1}
\vspace{0.1cm} \noindent
\textbf{Cross-domain Reliability of Embedding Space.}
% \subsubsection{Cross-domain Reliability of Embedding Space.}
We examine the geometry of multimodal embedding spaces across domains by extracting visual and textual CLIP features from image–caption pairs in the four PACS domains~\cite{dataset-pacs} (art painting, photo, cartoon, sketch). 
%embedding geometry of multimodal representations across domains.
% We extract visual and textual CLIP features from image-caption pairs within the four PACS~\cite{dataset-pacs} domains (art painting, photo, cartoon, sketch).
To obtain textual embeddings as rich as their visual counterparts,
% In order to extract equally informative textual embeddings to the visual counterpart, 
we generate captions with semantic and stylistic details (App.~\ref{sec:app:caption}), 
instead of using predefined prompts like \texttt{"a [domain] of [class]"},
% of using the name-only format like \texttt{`a [domain] of [class]'}, 
and feed them into a frozen CLIP text encoder.
This provides rich instance-level descriptions and enables a fair comparison between modalities.
% This design ensures that textual inputs contain rich instance-level descriptions, enabling a fair comparison between the two modalities.

%\joonseok{From the images in} the four PACS~\cite{dataset-pacs} domains (art painting, photo, cartoon, sketch), we extract CLIP visual features together with corresponding caption-based text features, and visualize them using t-SNE.
%Importantly, rather than adopting a standard prompt, \texttt{`a photo of [class]'}, or domain-conditioned prompts (\eg{\texttt{`a [domain] of [class]'}}), 
%we generate image-conditioned captions (see \cref{sec:app:caption}), which include both semantic and stylistic details, and feed them into the frozen CLIP text encoder.
%This design ensures that textual inputs contain rich instance-level descriptions, enabling a fair comparison between modalities: both visual and textual embeddings are allowed to encode comparably detailed content.

As shown in~\cref{fig:motiv}(a), the visual space (top) exhibits pronounced domain-dependent dispersion.
Even for the same class (\emph{e.g.}, person, highlighted with black boxes), the corresponding clusters significantly shift across domains.
In contrast, text embeddings (bottom) remain relatively stable,
%grouped by class, 
largely insensitive to stylistic variations present in the captions.
This contrast reveals a notable gap:
%an asymmetry:
while visual encoders capture fine-grained details including domain-specific cues, text embeddings remain %predominantly 
semantically structured.
%\red{class-centered}, 

%\subsection{Quantifying Modality Asymmetry}
%\label{sec:3-2}
\vspace{0.1cm} \noindent
\textbf{Encoded Information in Embedding Spaces.}
The previous experiment implies that the textual embedding space likely encode more domain-invariant information compared to the visual counterpart.
Also, one might wonder whether this behavior is specific to CLIP.
To answer these questions,
%To examine its generality,
we report in \cref{fig:motiv}(b) the normalized mutual information between the learned embeddings
and both class labels ($x$-axis) and domain labels ($y$-axis) across multiple encoders.

Across visual backbones (ViT~\cite{dosovitskiy2020vit}, CLIP-Image~\cite{pmlr-v139-radford21a}, and DINOv2~\cite{oquab2023dinov2}), their embeddings consistently encode substantial domain information $I(Z;\text{domain})$ alongside class information $I(Z;\text{class})$.
In contrast, language models (MPNet~\cite{10.5555/3495724.3497138}, MiniLM~\cite{10.5555/3495724.3496209}, and CLIP-Text~\cite{pmlr-v139-radford21a}) exhibit near-zero domain information, while primarily encoding class semantics.
This result strongly implies that domain entanglement is indeed inherent to expressive visual representations.
Preserving fine-grained variations, visual encoders inadvertently keep some domain-specific factors as well.
In contrast, textual representations exhibit much less stylistic variation, leading them to align more consistently with semantic structure.
Considering that domain generalization strictly requires isolating domain-invariant semantics from such variations, a natural hypothesis from this experiment is that textual space would be more appropriate to guide DG models.

%\subsection{Learning Dynamics under Domain-Dependent Guidance}
%\label{sec:3-3}
\vspace{0.1cm} \noindent
\textbf{Learning Dynamics under Domain-Dependent Guidance.}
A natural next question is if the guidance signals contain domain-dependent variations, how does this actually affect learning dynamics?
To investigate this, we estimate the local Lipschitz constant, approximated by the norm of the gradient of learned features with respect to unseen target inputs~\cite{Yang_2025_CVPR}.
This metric measures how sensitively the representation reacts to small perturbations under domain shift.

As shown in~\cref{fig:motiv}(c), student models (ViT~\cite{dosovitskiy2020vit} and ResNet-50~\cite{he2016deep})
distilled with visual signals
%trained with a standard classification loss and additional CLIP-based visual distillation
exhibit consistently higher Lipschitz values than those trained with textual guidance, indicating more input-sensitive mappings.
This result verifies our earlier hypothesis that visual guidance would expose the student to cross-domain conflicting cues (\cref{fig:motiv}(a)).
Fitting to such inconsistent signals makes the representation more sensitive to small input changes and compromises stability.
In contrast, textual guidance mitigates cross-domain conflict, yielding smoother and more stable representations under domain shift.

\subsection{Information-theoretic Perspective}
%\subsection{Theoretical Justification \eunyi{of Design Choices}}
\label{sec:motivation:theory}
Beyond our empirical diagnostics, we further provide an information-theoretic perspective on our two design choices: removing domain-contaminated visual guidance and introducing an explicit information bottleneck.
%theoretically justify our two core design choices, 
% \joonseok{from the information-theoretic perspective}.
We interpret this through the lens of finite model capacity (formal assumptions and derivations are provided in App.~\ref{sec:app:img_guidance}).

An input image $X$ can be decomposed into domain-invariant and domain-specific (spurious) components: $X = (X_{\text{inv}}, X_{\text{sp}})$.
Denoting the output of the student model $\phi$ as $S = \phi(X)$ with finite capacity $C$, we have $I(S;X_\text{inv}) + I(S;X_\text{sp}) \leq C$.
% The finite capacity $C$ induces
%Let a student model $S$ with finite capacity $C$ satisfy $I(S;X_\text{inv}) + I(S;X_\text{sp}) \leq C$, inducing 
This constraint implies a trade-off between invariant semantics and spurious domain styles.
Because visual guidance depends on the input image, 
it carries domain-specific variation $X_\text{sp}$ into the supervision signal.
% it propagates $X_\text{sp}$ through the supervision signal.
Learning from such guidance increases $I(S;X_\text{sp})$, thereby reducing the capacity available for $X_\text{inv}$.
Since labels depend only on $X_\text{inv}$, this limits the attainable predictive information and degrades generalization to unseen domains.
This provides additional theoretical support for removing visual guidance.

However, the removal alone does not fully resolve the capacity allocation problem.
% Although it prevents domain signals $X_\text{sp}$ from entering supervision, 
Although domain signals $X_\text{sp}$ are no longer injected through supervision,
they are still present in the input and may be encoded by the model during training.
In practice, a high-capacity model can learn separate domain-specific feature pathways that reach the same semantic target, leaving the representation entangled with $X_\text{sp}$.
% To quickly minimize alignment loss, a high-capacity model can still establish multiple domain-specific routes to the same text anchor, leaving the learned representation 
To explicitly restrict this spurious capacity allocation, we introduce a Text-Anchored Information Bottleneck in the next section.

% Beyond empirical instability, why does domain-contaminated visual guidance fundamentally restrict what the model can learn? 
% We provide an information-theoretic analysis showing that when supervision entangles semantics with spurious domain factors, invariant learning becomes structurally constrained under finite model capacity.

% Let the guidance signal be a mixture of visual and textual representations, controlled by an image-guidance weight $\epsilon$.
% Textual guidance is approximately domain-invariant, $I(G_t;X_\text{sp})\approx0$, where $X_\text{sp}$ denotes spurious domain factors.
% In contrast, visual guidance inherently encodes such variation, yielding $I(G_i;X_\text{sp})>0$.

% For a student representation $S$ with finite capacity $C$, incorporating domain-contaminated guidance necessarily allocates capacity to spurious information. 
% As reliance on visual guidance increases (larger $\epsilon$),
% $I(S;X_\text{sp})$ increases.
% Under capacity constraints, this reduces the attainable mutual information with true labels $I(S;Y)$, leading to degraded generalization:
% \begin{equation}
% \epsilon\uparrow  \text{ }\rightarrow\text{ } I(S;X_\text{sp})\uparrow \text{ }\rightarrow\text{ } I(S;Y)\downarrow \text{ }\rightarrow\text{ } \mathcal{E}_\text{test}\uparrow,
% \end{equation}
% with derivations provided in App.~\ref{sec:app:img_guidance}.

\section{Method}
\label{sec:4-method}
% https://docs.google.com/presentation/d/1R6CGa3xkgWLrSoNvPaLKou8v4iTYgqGhoft0sMkxiH4/edit?slide=id.g3bd849e84a3_0_4#slide=id.g3bd849e84a3_0_4

\begin{figure}[t]
  \centering
  \includegraphics[width=\linewidth]{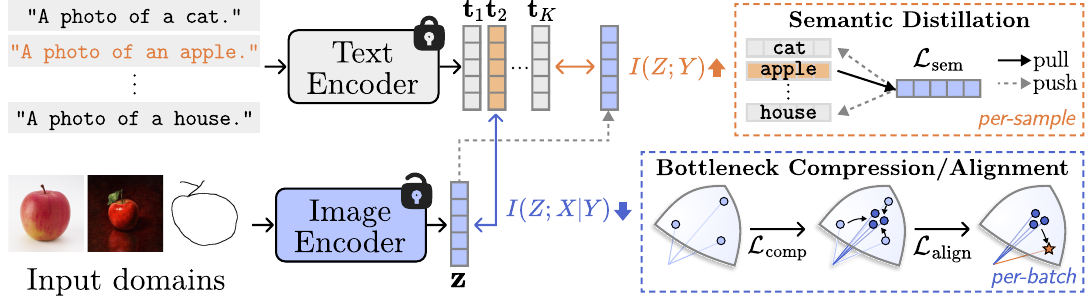}
  \caption{\textbf{Overview of Text-Anchored Information Bottleneck}. Text guidance is the primary source of domain-invariance under our %by optimizing the 
  Conditional Entropy Bottleneck (CEB) formulation, composed of two parts: i) Semantic distillation ($\mathcal{L}_\text{sem}$) maximizes $I(Z;Y)$ by pulling image representations toward text anchors, and ii) Bottleneck compression and alignment minimizes $I(Z;X|Y)$ to suppress domain-specific variations, achieved by encouraging intra-class concentration via $\mathcal{L}_\text{comp}$ and aligning class-wise mean feature with text anchors via $\mathcal{L}_\text{align}$.}
  \label{fig:arch}
  \vspace{-0.4cm}
\end{figure}

Our previous diagnostics verify that domain invariance is largely determined by the structure of supervision.
Instead of relying on visually entangled guidance, which inevitably propagates spurious domain variations, we propose to anchor the training entirely to fixed, domain-invariant text embeddings, namely, Text-Anchored Information Bottleneck, illustrated in \cref{fig:arch}.
We first formalize the domain generalization problem and introduce the Information Bottleneck formulations (\cref{sec:4-1}), then present our text-anchored framework with two mechanisms: preserving semantic structure in the text space and suppressing spurious visual cues (\cref{sec:4-2}).
% Beginning with formalizing the domain generalization problem(\cref{sec:4-1}), we introduce our text-anchored framework with two mechanisms: to preserve semantic structures in text space, and to explicitly drop spurious visual cues (\cref{sec:4-2}).

\subsection{Preliminary}
\label{sec:4-1}
We first formulate the DG problem and introduce the Information Bottleneck and its conditional variant, which governs our semantic purification strategy.

\vspace{0.1cm} \noindent
\textbf{Problem Formulation.}
In domain generalization, training data from $K$ source domains $\mathcal{D}_S = \{\mathcal{D}_1, \mathcal{D}_2, \dots, \mathcal{D}_K\}$ are given, where $\mathcal{D}_k = \{(x_i^{(k)}, y_i^{(k)})\}_{i=1}^{N_k}$ from each domain $k$ consists of $N_k$ samples and follows a probability distribution $P_k(X, Y)$.
The core challenge is distribution shift, where $P(Y|X)$ remains invariant while the marginals differ across domains ($P_i(X)\neq P_j(X)$).
The task aims to learn a feature encoder $Z=f_\theta(X)$ that generalizes to an unseen target domain $\mathcal{D}_T$, where $\mathcal{D}_T\cap\mathcal{D}_S=\emptyset$.
% In the standard DG, data from $K$ source domains $\mathcal{D}_S = \{\mathcal{D}_1, \mathcal{D}_2, \dots, \mathcal{D}_K\}$ are utilized, 
% where each domain $\mathcal{D}_k = \{(x_i^{(k)}, y_i^{(k)})\}_{i=1}^{N_k}$ follows a distribution $P_k(X, Y)$ over \red{$\mathcal{X} \times \mathcal{Y}$}, \joonseok{and $N_k$ is the number of samples in domain $k$}. 
% The core challenge is distribution shift: while $P(Y|X)$ remains invariant, the marginal distributions differ across domains; that is, $P_i(X) \neq P_j(X)$ for $i \neq j$.
% Our objective is to learn a robust feature encoder $Z = f_{\theta}(X)$, where $f_\theta : \mathcal{X} \to \red{\mathcal{Z}}$, that generalizes to a completely unseen target domain $\mathcal{D}_T$, where $\mathcal{D}_T \cap \mathcal{D}_S = \emptyset$.

\vspace{0.1cm} \noindent
\textbf{Information Bottleneck.}
To learn robust representations $Z$, we adopt the Information Bottleneck (IB)~\cite{tishby99information}, which seeks features that are maximally predictive of labels $Y$ while compressing the input $X$: \begin{equation}
\mathcal{L}_\text{IB} = -I(Z;Y) + \beta I(Z;X),
\label{eq:ib}
\end{equation}
where $I(\cdot;\cdot)$ denotes mutual information and
$\beta > 0$ controls the trade-off between prediction and compression.
Assuming the Markov chain $Y\leftrightarrow X\leftrightarrow Z$, IB encourages $Z$ to retain label-relevant information while discarding irrelevant variations in $X$.

Minimizing $I(Z;X)$ alone, however, does not explicitly distinguish label-relevant semantics from domain-specific factors, since domain cues may still be predictive of $Y$ within the training domains.
Therefore, we minimize the Conditional Information Bottleneck (CEB)~\cite{Fischer_2020}:
\begin{equation}
    \mathcal{L}_\text{CEB} = -{I(Z;Y)+\beta I(Z;X|Y)},
    \label{eq:ceb}
\end{equation}
which replaces $I(Z;X)$ with $I(Z;X|Y)$.
By conditioning on the label $Y$, CEB removes input information that is unnecessary given $Y$. 
This provides a direct mechanism for suppressing spurious domain-specific variations, as labels serve as semantic anchors that isolate task-relevant information.

\subsection{Text-Anchored Information Bottleneck}
\label{sec:4-2}
\cref{fig:arch} presents our architecture-agnostic training framework for optimizing the CEB under fixed textual semantics. 
With $K$ classes, we obtain semantic anchors $T = [\mathbf{t}_1, …, \mathbf{t}_K ]^\top \in \mathbb{R}^{K \times d}$, where each $\mathbf{t}_k \in \mathbb{R}^d$ is the frozen CLIP text embedding of the prompt \texttt{`a photo of a [class]'}. 
Although we have used rich captions in \cref{sec:3-motivation} to fairly match image richness, here we adopt the simplest prompt, which strips instance-level noise while keeping class identity.
A trainable visual encoder $f$ maps an input image $\boldsymbol{x}$ to $\boldsymbol{z}=f_\theta(\boldsymbol{x}) \in \mathbb{R}^d$.
Motivated by~\cref{sec:3-motivation}, we treat these fixed text embeddings as the primary source of domain invariance, explicitly anchoring the representation space to domain-stable semantics rather than relying on invariance to emerge implicitly or using text as auxiliary guidance.
%Under this design, the CEB is instantiated as:
%\begin{equation}
%    \mathcal{L}_\text{CEB} = -{I(Z;Y)+\beta I(Z;X\mid Y)},
%    \label{eq:ceb}
%\end{equation}
%where conditioning corresponds to stable semantic anchors.
We derive concrete forms of the two complementary objectives, maximizing predictive sufficiency $I(Z;Y)$ and minimizing anchor-inconsistent variation $I(Z;X|Y)$, comprising our CEB formulation in \cref{eq:ceb}.
This yields three terms: a semantic distillation loss $\mathcal{L}_\text{sem}$ that maximizes $I(Z;Y)$, and compression and alignment losses $\mathcal{L}_\text{comp}$, $\mathcal{L}_\text{align}$ that minimize $I(Z;X|Y$), together forming our final objective (\cref{eq:loss_full}).
We further detail each below.
% This formulation yields two complementary objectives—maximizing predictive sufficiency and minimizing anchor-inconsistent variation—which we derive below.

\vspace{0.1cm} \noindent
\textbf{Maximizing $I(Z;Y)$.}
The mutual information term $I(Z; Y)$ in \cref{eq:ceb} quantifies the dependency between the representation $Z$ and the target $Y$, defined as
\begin{equation}
  I(Z; Y) = \mathbb{E}_{Z, Y} \left[ \log \frac{p(Z,Y)}{p(Z)p(Y)} \right] 
    = \mathbb{E}_{Z, Y} \Big[\log p(Y|Z)\Big] + H(Y),
\end{equation}
where $H(Y)$ is constant.
Thus, maximizing $I(Z; Y)$ is equivalent to maximizing $\mathbb{E}_{Z, Y} [\log p(Y|Z)]$, \emph{i.e.}, minimizing the cross-entropy loss~\cite{alemi2017deep}.
We parameterize $p(Y|Z)$ using fixed text embeddings as class prototypes.
For a sample with label $k$, we define:
\begin{equation}
  \mathcal{L}_\text{sem} = -\log \frac{\exp(\cos(\boldsymbol{z}, \mathbf{t}_k)/\tau)}{\sum_{k' \in K} \exp(\cos(\boldsymbol{z}, \mathbf{t}_{k'})/\tau)},
\end{equation}
where $\mathbf{t}_k$ is the text embedding of class $k$ and $\tau$ is a temperature.
% This corresponds to a prototype-based cross-entropy objective.
% This objective pulls representations toward their class anchors and pushes them away from others, organizing the space according to the fixed textual semantic structure (\cref{fig:arch}, upper right).
This objective pulls representations toward their class anchors and pushes them away from others, thereby distilling the semantic structure of the textual space into the learned representation (\cref{fig:arch}, upper right).
% By doing so, it is expected to distill the semantic structure of the textual space into the learned representation.
% Additionally, it can be interpreted as an InfoNCE loss~\cite{oord2019representationlearningcontrastivepredictive}, reinforcing its connection to contrastive learning, where ($\mathbf{z}, \mathbf{t}_k$) forms the positive pair, while other class embeddings serve as negatives, promoting directional consistency and encouraging image features to reflect the semantic structure of the textual embedding space.

\vspace{0.1cm} \noindent
\textbf{Minimizing $I(Z;X|Y)$.}
We minimize the second term $I(Z;X|Y)$ in \cref{eq:ceb} to suppress spurious domain variations.
We first derive its variational upper bound, and then convert it into a tractable training objective.

% 1. UB Derimation
Starting from the definition of conditional mutual information,
\begin{equation}
I(Z;X|Y) \coloneqq \mathbb{E}\left[\log \frac{p(X,Z| Y)}{p(X| Y)p(Z| Y)}\right] = \mathbb{E}\left[\log \frac{p(Z| X,Y)}{p(Z| Y)}\right] = \mathbb{E}\left[\log \frac{p(Z| X)}{p(Z| Y)}\right],
\end{equation}
where the last equality follows from the Markov chain $Y\leftrightarrow X \leftrightarrow Z$ assumption, implying $p(Z|X, Y) = p(Z|X)$.
Since the true marginal $p(Z|Y)$ is intractable, we introduce a variational approximation $q(Z|Y)$.
% , defined by the fixed text embeddings serving as semantic anchors.
By non-negativity of KL divergence,
\begin{equation} 
  I(Z;X|Y) \le \mathbb{E}[\log p(Z|X)]-\mathbb{E}[\log q(Z|Y)]
  = \mathbb{E}\big [\mathrm{KL} (p(Z|X)\|q(Z|Y)) \big].
  \label{eq:ub}
\end{equation}

% 2. Objective Approximation
We now derive a stable training objective from this bound.
For samples belonging to class $k$, \cref{eq:ub} becomes
\begin{equation}
  \mathbb{E}_{X|y=k}\!\big[\mathrm{KL} (p(Z|X)\|q(Z|y=k))\big]=
  \mathbb{E}_{X|y=k}\big[\mathbb{E}_{Z}\big[\log p(Z|X)-\log q(Z|y=k)\big]\big],
  \label{eq:kl_2ndterm}
\end{equation}

Since CLIP embeddings are $l_2$-normalized and lie on the unit hypersphere, we model both $p$ and $q$ as von Mises–Fisher (vMF) distributions \cite{fisher1953dispersion}, parameterized by a mean direction $\boldsymbol{\mu}$ and concentration $\kappa$.
A vMF can be viewed as the spherical analogue of a Gaussian, concentrating probability mass around a direction on the unit sphere.
% Intuitively, a vMF is
Specifically, $q(Z|y=k)$ is defined as a vMF centered at the fixed text embedding $\mathbf{t}_k$ obtained from the frozen text encoder, with a fixed concentration $\kappa_q$, while $p(Z|x)$ is centered at the image feature $\boldsymbol{z}=f_\theta(x)$, with concentration $\kappa_p$.
% For a class $k$, the prior $q(Z|y=k)$ is defined as a vMF centered at the fixed text anchor $\mathbf{t}_k$ with a constant $\kappa > 0$.
% For a data sample \eunyi{$x$}, the encoder models \eunyi{$p(Z|x)$} as a highly concentrated vMF centered at $\boldsymbol{z}$, the image feature encoded by the frozen image encoder.
% Approximating this with an empirical expectation over data samples gives:
With $\kappa_p$ fixed, the first term in \cref{eq:kl_2ndterm}, $\mathbb{E}_Z[\log p(Z|X)]$, becomes constant with respect to $\theta$. 
Thus, the optimization reduces to minimizing the second term, $-\mathbb{E}_{X|y=k}\mathbb{E}_Z[\log q(Z|y=k)]$.

Approximating $Z\sim p(Z|x)$ by its mean feature $\boldsymbol{z}=f_\theta(x)$ (with a large $\kappa_p$), and using the vMF density $q(\boldsymbol{z}|y=k)=C_d(\kappa_q)\exp{(\kappa_q\boldsymbol{t}_k^\top\boldsymbol{z})}$, the log-density can be substituted into \cref{eq:kl_2ndterm}. 
Applying the empirical expectation over a minibatch $\mathcal{B}_k$ yields
\begin{equation}
  -\frac{1}{|\mathcal{B}_k|}\sum_{i\in \mathcal{B}_k}[\log C_d(\kappa_q)+ \kappa_q\mathbf{t}_k^\top \boldsymbol{z}_i]
  = \text{const}- \kappa_q\, \mathbf{t}_k^\top\left(\frac{1}{|\mathcal{B}_k|}\sum_{i\in \mathcal{B}_k}\boldsymbol{z}_i\right),
\end{equation}
where both $\kappa_q$ and $C_d(\kappa_q)$ are constants.
Let $\bar{\boldsymbol{z}}_k =\sum_{i\in \mathcal{B}_k}\boldsymbol{z}_i/|\mathcal{B}_k|$ denote the mean feature of the class $k$, % within a minibatch.
the objective reduces to
\begin{equation}
  -\boldsymbol t_k^\top \bar{\boldsymbol z}_k =-\,\|\boldsymbol{t}_k\|\|\bar{\boldsymbol z}_k\|\cos(\boldsymbol t_k,\bar{\boldsymbol{z}}_k)= -\,\|\bar{\boldsymbol{z}}_k\|\cos(\boldsymbol t_k,\bar{\boldsymbol{z}}_k),
\end{equation}
since $\|\boldsymbol{t}_k\|=1$.
% In practice, $\bar{\boldsymbol{z}}_k$ is estimated within a mini-batch.
Empirically, however, directly optimizing this multiplicative form couples the magnitude and the cosine terms, which can lead to poorly conditioned gradients during training (\emph{e.g.}, the cosine term may quickly saturate, weakening gradients acting on the $\|\bar{\boldsymbol{z}}\|$ term). 
We therefore adopt a separable surrogate objective that independently encourages (i) a large intra-class resultant length $\|\bar{\boldsymbol{z}}_k\|$, and (ii) strong directional alignment $\cos(\mathbf t_k,\bar{\boldsymbol{z}}_k)$ with the text anchor $\mathbf t_k$, to stabilize optimization.
% \red{This derivation requires jointly maximizing the magnitude of the class mean $\|\bar{\boldsymbol{z}}_k\|$ and its directional similarity $\cos(\mathbf t_k,\bar{\boldsymbol{z}}_k)$ with the corresponding text anchor $\mathbf{t}_k$.}
From the magnitude term, we define:
\begin{equation}
  \mathcal{L}_{\mathrm{comp}} = -\frac{1}{|K_{\mathcal B}|}\sum_{k \in K_{\mathcal B}}\|\bar{\boldsymbol{z}}_k\|,
  \label{eq:loss_comp}
\end{equation}
where $K_\mathcal{B}$ denotes the set of classes present in a mini-batch $\mathcal{B}$.
This encourages features of the same class to concentrate along a coherent direction (\cref{fig:arch}, lower right).
From the cosine term, we define
\begin{equation}
  \mathcal{L}_{\mathrm{align}} = -\frac{1}{|K_{\mathcal B}|}\sum_{k \in K_{\mathcal B}}\cos\! \left(\mathbf t_k, 
  \frac{\bar{\boldsymbol{z}}_k}{\|\bar{\boldsymbol{z}}_k\|} \right)
  \label{eq:loss_align}
\end{equation}
which aligns each class mean with its text anchor. The final objective is
\begin{equation}
  \mathcal{L}=\mathcal{L}_{\mathrm{sem}}+\beta_1 \mathcal{L}_{\mathrm{align}}+\beta_2 \mathcal{L}_{\mathrm{comp}},
  \label{eq:loss_full}
\end{equation}
where $\beta_1$ and $\beta_2$ balance the two effects.

Unlike prior IB-based DG methods~\cite{Yu2023INSUREAI,10.1007/978-3-030-58607-2_12,Li2021InvariantIB} that learn explicit Gaussian parameters (\emph{e.g.}, $\boldsymbol{\mu}$, $\boldsymbol{\sigma}$) from each image to match a simple unconditional prior (\emph{e.g.}, $\mathcal{N}(0, \mathbf{I})$), irrespective of class semantics, we impose a class-conditional prior anchored by fixed text embeddings, enabling more reliable compression of the representations.
% \red{[see comment]}
\section{Experiments}
\label{sec:5-results}

\subsection{Experimental Setting}

\textbf{Datasets.}
We evaluate on six standard DG benchmarks: TerraIncognita~\cite{dataset-ti}, OfficeHome~\cite{dataset-oh}, VLCS~\cite{dataset-vlcs}, PACS~\cite{dataset-pacs}, DomainNet~\cite{dataset-dn}, and NICO\textsuperscript{++}~\cite{dataset-nico}.
The first four contain four domains each with 10, 65, 5, and 7 classes, respectively. For large-scale evaluation, we use the latter two, DomainNet (6 domains, 345 classes) and NICO\textsuperscript{++} (6 domains, 60 classes).

\vspace{0.1cm} \noindent
\textbf{Baselines.}
% We compare against representative DG methods across diverse backbones and pretraining schemes, covering classical DG, multimodal distillation, and CLIP-specialized adaptation.
% Linear Probing (LP) trains a linear classifier on a frozen backbone, assessing the inherent cross-domain robustness of pretrained features.
% MIRO~\cite{10.1007/978-3-031-20050-2_26} represents classical DG without text guidance.
% RISE~\cite{Huang2023ASS} and VL2V~\cite{Addepalli_2024_CVPR} rely on visaul distillation from a teacher image encoder, utilizing text embeddings as auxiliary targets for simple cross-modal alignment.
% CLIPood~\cite{pmlr-v202-shu23a} is a CLIP-specific method employing margin-based softmax and a beta-moving average to preserve zero-shot weights.
% We compare our proposed approach with representative DG methods across diverse backbones using several key baselines.
Across diverse backbones, representative DG methods vary by architecture. For consistent comparison, we focus on a set of key baselines.
Linear probing (LP) evaluates frozen features, while MIRO~\cite{10.1007/978-3-031-20050-2_26} represents classical DG without external guidance. 
% RISE~\cite{Huang2023ASS} and VL2V~\cite{Addepalli_2024_CVPR} utilize cross-modal distillation, using an CLIP image encoder as the primary teacher and simple alignment objectives for text supervision. 
% For CLIP-dedicated methods such as CLIPood~\cite{pmlr-v202-shu23a}, zero-shot preservation retains original CLIP visual features on trained weights, effectively maintaining CLIP visual guidance. On non-CLIP backbones, preservation instead applies to backbone-specific weights, and CLIP visual features are no longer part of the adaptation process.
RISE~\cite{Huang2023ASS} and VL2V~\cite{Addepalli_2024_CVPR} perform cross-modal distillation, explicitly using a CLIP image encoder as the primary teacher with simple alignment objectives for text supervision. In contrast, CLIP-dedicated methods such as CLIPood~\cite{pmlr-v202-shu23a} do not employ an external teacher but implicitly retain original CLIP visual features through zero-shot preservation. When applied to non-CLIP backbones, however, this preservation operates on backbone-specific weights, so CLIP visual features are no longer involved in adaptation and guidance is provided only by CLIP text embeddings.

\vspace{0.1cm} \noindent
\textbf{Implementation Details.}
We follow the DomainBed~\cite{gulrajani2021in} protocol. All datasets use the standard leave-one-domain-out setting, where one domain is held out for testing while training on the remaining domains. For NICO\textsuperscript{++}, we adopt a leave-one-group-out protocol, where multiple domains are grouped and one entire group is reserved for evaluation.
Model selection is based on validation accuracy using a 20\% split of training data. 
We use $\beta_1$ = 0.1 and $\beta_2$ = 1.0 based on ResNet-50 on PACS, and apply them across all datasets and backbones (sensitivity in App.~\ref{sec:app:hp_sensitivity}). 
Results are averaged over three runs, with standard deviations reported in App.~\ref{sec:app:std}.
See App.~\ref{sec:app:exp_details} for more details.

\subsection{Comparison with Baselines}
\label{sec:exp:main_comparison}

\begin{table}[t]
\centering
\renewcommand{\arraystretch}{0.94}
% \footnotesize
\caption{
  \textbf{DG performance comparison across representative backbones}. The ‘G’ column denotes CLIP visual (V) or textual (T) guidance. The \textbf{best} and \underline{second-best} results are highlighted.
  %the total average is reported (Avg).
  * denotes our implementation.}
%\vspace{-0.3cm}
\setlength{\tabcolsep}{6pt}
\resizebox{\linewidth}{!}{
\begin{tabular}{l|c|*5{w{c}{1.6cm}}|c}
\toprule
\multicolumn{1}{c|}{Method} & G & VLCS & PACS & OfficeHome & TerraInc & DomainNet & Average  \\ 
\midrule[1pt]
\rowcolor{gray!10}
\multicolumn{8}{c}{\textit{ResNet-50 pretrained on ImageNet-1k}}\\ 
LP&-& 78.1          & 86.2          & 68.4        & 46.3        & 41.2 &  64.0       \\
MIRO~\cite{10.1007/978-3-031-20050-2_26}& -&79.0          & 85.4          & 70.5        & 50.4        & 44.3 &  65.9  \\
SAGM~\cite{Wang2023SharpnessAwareGM}&-& 80.0          & 86.6          & 70.1        & 48.8        & 45.0 &  66.1  \\
GESTUR~\cite{Lew2023GradientEF}& -&80.1          & 88.0 & 71.1        & 51.3 & 46.3  & 67.4  \\
INSURE~\cite{Yu2023INSUREAI}&-& -             & 89.3 & 72.0        & 53.1 & \underline{48.0}  &  -        \\
CLIPood*\cite{pmlr-v202-shu23a}& T&76.7          & 88.8          & 70.3        & 44.7        & - & -         \\
RISE~\cite{Huang2023ASS}& V,T& \textbf{81.7}          & \underline{89.4}          & 71.6        & 52.3        & 46.5 & \underline{68.3}         \\
VL2V~\cite{Addepalli_2024_CVPR}& V,T& 79.2 & 86.7          & \underline{74.4} & \underline{53.5}        & 47.7 & \underline{68.3} \\
\rowcolor{cyan!10}
Ours & T&\textbf{81.7} & \textbf{96.9} & \textbf{79.0} & \textbf{59.9} & \textbf{58.3} & \textbf{75.4} \\ 
\midrule[1pt]
\rowcolor{gray!10}
\multicolumn{8}{c}{\textit{ViT-B/16 pretrained on ImageNet-1k}}\\ 
LP& -& 79.5          & 81.5          & 82.8        & 42.2        & 50.5 & 67.3 \\
MIRO*\cite{10.1007/978-3-031-20050-2_26}& -&80.4          & 81.5          & 74.9        & 44.5        & - & -         \\
CLIPood*\cite{pmlr-v202-shu23a}& T&80.4          & 87.1          & 80.9        & 44.3        & - & -         \\
RISE*\cite{Huang2023ASS}& V,T&\underline{84.2}          & 91.0          & 80.3        & 44.6 & 56.6 & 71.3 \\
VL2V~\cite{Addepalli_2024_CVPR}& V,T&81.9 & \textbf{94.9}          & \underline{85.7} & \underline{55.4}        & \underline{59.4} & \underline{75.5} \\
\rowcolor{cyan!10}
Ours & T&\textbf{86.2} & \underline{94.1} & \textbf{86.4} & \textbf{62.2} & \textbf{68.8} & \textbf{79.5}  \\ 
\midrule[1pt]
\rowcolor{gray!10}
\multicolumn{8}{c}{\textit{CLIP-ViT-B/16 pretrained on private dataset (400M)}}\\ 
% \midrule
LP&-& 83.4          & 97.2 & 82.3        & 57.3        & 58.2 & 75.7 \\
CLIP-ZS&-& 82.4 & 96.1 & 82.3 & 34.4 & 49.7 & 69.0 \\
MIRO~\cite{10.1007/978-3-031-20050-2_26}&-& 82.2          & 95.6          & 82.5        & 54.3     & 54.0 & 73.7 \\
GESTUR~\cite{Lew2023GradientEF}& -&82.8          & 96.0          & 84.2        & 55.7        & 58.9 & 75.5   \\
CAR-FT~\cite{10.1007/s11263-023-01951-2}&V,T& 85.5 & 96.8          & 85.7        & 61.9 & 62.5 & 78.5         \\
CLIPood~\cite{pmlr-v202-shu23a}& V,T&85.0          & 97.3 & 87.0        & 60.4        & 63.5 & 78.6         \\
CLIPCEIL~\cite{yu2024clipceil}& V,T&85.2          & 97.2          & \underline{87.7} & 62.0 & \underline{63.6} & \underline{79.1} \\
CLIP-DPR~\cite{Cheng2024DisentangledPR}   & V,T&\underline{86.4}          & \underline{97.5}          & 86.1 & 57.1 & 62.1 & 77.8 \\
CLIP-DTP~\cite{CLIP-DTP} & V,T&84.8 & 97.0  & \underline{87.7} & \underline{63.3} & 63.1 & 79.2 \\
RISE~\cite{Huang2023ASS}& V,T&80.6          & 93.3          & 78.4        & 49.6        & 55.4 & 71.5 \\
VL2V~\cite{Addepalli_2024_CVPR}& V,T&83.3          & 96.7          & 87.4        & 58.5        & 62.8 & 77.7 \\
\rowcolor{cyan!10}
Ours & T&\textbf{89.0} &\textbf{98.5}  & \textbf{93.2} & \textbf{75.1} & \textbf{75.8} & \textbf{86.3} \\ 
\bottomrule
\end{tabular}}
\label{tab:main}
\end{table}

% \begin{table*}[t]
% \centering
% \caption{
% \textbf{NICO++ DG performance.}.
% }
% \label{tab:exp1}

% \scriptsize
% \setlength{\tabcolsep}{2pt}  % column spacing

% \begin{subtable}[t]{0.48\textwidth}
% \centering
% \caption{ResNet-50 (ImageNet-1k)}
% \begin{tabular}{l|cc|cc|cc|c}
% \toprule
% Method & A & R & D & G & O & W & Avg \\
% \midrule
% ERM & 85.3 & 85.6&78.6&	86.5&	82.0&	76.7&	82.5 \\
% MIRO &  82.3&81.2&73.5&81.5&77.9&72.0&78.1 \\
% VL2V & 85.2&84.3&77.1&87.1&82.7&78.9&82.6\\
% Ours & \textbf{95.7}&\textbf{97.3}&\textbf{94.4}&\textbf{97.7}&\textbf{96.4}&\textbf{92.9}&\textbf{95.7}\\
% \bottomrule
% \end{tabular}
% \end{subtable}
% \hfill
% %
% \begin{subtable}[t]{0.48\textwidth}
% \centering
% \caption{ViT-B/16 (CLIP)}
% \begin{tabular}{l|cc|cc|cc|c}
% \toprule
% Method & A & R & D & G & O & W & Avg \\
% % \hline
% \midrule
% ERM & 91.4&92.2&89.8&92.7&90.4&84.6&90.2 \\
% MIRO & 92.1&91.4&86.7&92.3&89.3&83.8&89.3 \\
% SRE & 91.4&92.3&90.4&93.2&90.8&86.4&90.8 \\
% CLIPood & 93.0&94.3&89.6&93.7&92.0&86.8&91.5 \\
% VL2V &92.7&92.8&90.1&94.7&92.3&88.5&91.8 \\
% Ours & \textbf{97.9}&\textbf{98.5}&\textbf{97.7}&\textbf{98.7}&\textbf{98.3}&\textbf{94.0}&\textbf{97.5} \\
% \bottomrule
% \end{tabular}
% \end{subtable}
% \label{tab:nico}
% \end{table*}

\begin{table*}[t]
\centering
\caption{\textbf{Performance on NICO\textsuperscript{++}.} Results for ResNet-50 and CLIP backbones evaluated via a leave-one-group-out protocol on predefined target pairs: Autumn \& Rock, Dim \& Grass, and Outdoor \& Water.
}
\label{tab:nico}
\scriptsize
\setlength{\tabcolsep}{3pt} % Tighten padding to fit all columns
\resizebox{\linewidth}{!}{
\begin{tabular}{l | cc|cc|cc |c | cc|cc|cc| c}
\toprule
& \multicolumn{7}{c|}{\cellcolor{gray!10}\textit{ResNet-50 pretrained on ImageNet-1k}} & \multicolumn{7}{c}{\cellcolor{gray!10}\textit{CLIP-ViT-B/16}} \\
Method & A & R & D & G & O & W & Avg & A & R & D & G & O & W & Avg \\
\midrule
LP     & 85.3 & 85.6 & 78.6 & 86.5 & 82.0 & 76.7 & 82.5 & 91.4 & 92.2 & 89.8 & 92.7 & 90.4 & 84.6 & 90.2 \\
MIRO*~\cite{10.1007/978-3-031-20050-2_26} & 82.3 & 81.2 & 73.5 & 81.5 & 77.9 & 72.0 & 78.1 & 92.1 & 91.4 & 86.7 & 92.3 & 89.3 & 83.8 & 89.3 \\
VL2V*~\cite{Addepalli_2024_CVPR} & 85.2 & 84.3 & 77.1 & 87.1 & 82.7 & 78.9 & 82.6 & 92.7 & 92.8 & 90.1 & 94.7 & 92.3 & 88.5 & 91.8 \\
SRE~\cite{Wang2025SimulateRA} & -    & -    & -    & -    & -    & -    & -    & 91.4 & 92.3 & 90.4 & 93.2 & 90.8 & 86.4 & 90.8 \\
CLIPood*~\cite{pmlr-v202-shu23a} & -    & -    & -    & -    & -    & -    & -    & 93.0 & 94.3 & 89.6 & 93.7 & 92.0 & 86.8 & 91.5 \\
\midrule
\rowcolor{cyan!10}
Ours    & \textbf{95.7} & \textbf{97.3} & \textbf{94.4} & \textbf{97.7} & \textbf{96.4} & \textbf{92.9} & \textbf{95.7} & \textbf{97.9} & \textbf{98.5} & \textbf{97.7} & \textbf{98.7} & \textbf{98.3} & \textbf{94.0} & \textbf{97.5} \\
\bottomrule
\end{tabular}}
\end{table*}

\textbf{Main Results.}
\cref{tab:main} shows that our method achieves the state-of-the-art performance across various standard backbones.
Existing approaches exhibit architectural bias: KD methods (\emph{e.g.}, RISE, VL2V) favor conventional encoders, while CLIP-specialized methods favor CLIP backbones, and both degrade when evaluated beyond their original setup.
In contrast, ours consistently improves the performance without backbone-specific design, demonstrating its universal robustness.

\vspace{0.1cm} \noindent
\textbf{Robustness to Background Shift.}
We further compare the performance of competing DG models in \cref{tab:nico} on NICO\textsuperscript{++}, which focuses on background-driven shift.
%Under background-driven shift on NICO\textsuperscript{++} (\cref{tab:nico}),
Notably, the margin becomes more pronounced; ResNet-50 reaches 95.7\% accuracy, nearly closing the gap to CLIP (97.5\%).
This suggests that this setting is particularly better-aligned with our text-anchored bottleneck, since background variations are more readily separable from foreground semantics than the style shifts in~\cref{tab:main}, where object appearance itself can vary.
In such cases, varying background is more likely to be filtered as spurious style.
We further discuss in App.~\ref{sec:app:failure} that this purification improves overall robustness, though it may fail in rare cases where contextual cues are genuinely required for identification.

\begin{figure}[tb]
  \centering
  \includegraphics[width=\linewidth]{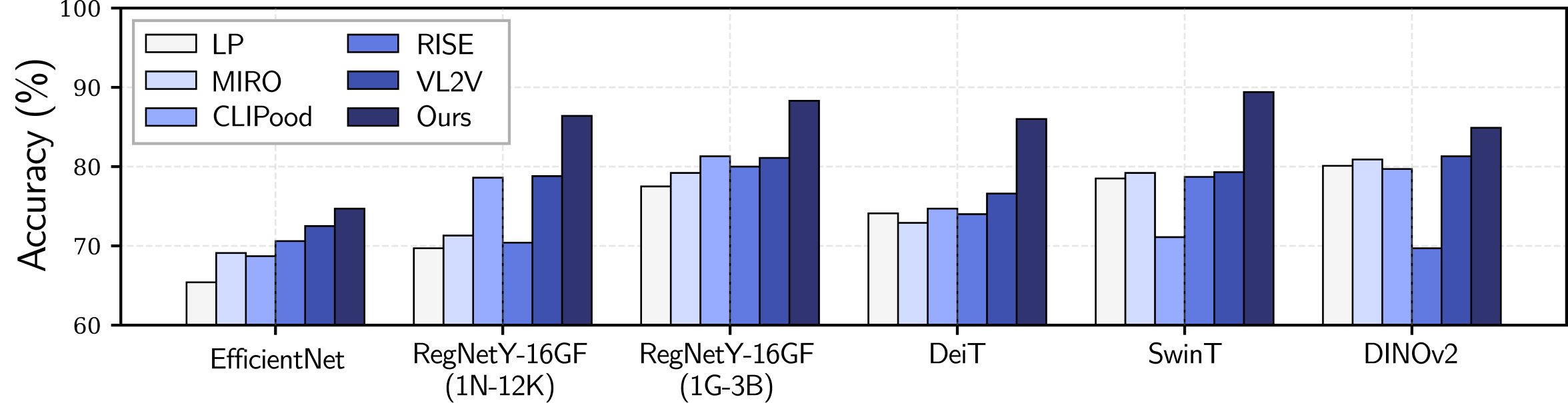}
  \caption{\textbf{DG performance across diverse backbones.} Average accuracy is reported (see App.~\ref{sec:app:backbones} for details). We evaluate across CNNs (EfficientNet, RegNetY) and Transformers with different architectures and pretraining schemes (SwinT, DeiT, DINOv2).}
  \label{fig:backbones}
\end{figure}

\vspace{0.1cm} \noindent
\textbf{Generalization across Backbones.}
While \cref{tab:main} evaluates standard architectures, \cref{fig:backbones} extends the comparison to a broader range of backbones, including lightweight models, CNNs, and diverse transformers.
Across all backbones, ours is the only method that consistently maintains a clear margin over LP.

Interestingly, when the pretrained backbone is already strong (\emph{i.e.}, high LP performance), other state-of-the-art methods often shrink to near-zero or even negative margins (\emph{e.g.}, MIRO on SwinT and RISE on DINOv2).
This implies that KD approaches (RISE, VL2V) risk distorting well-structured semantic spaces, while regularization-based methods (MIRO, CLIPood) largely preserve them without actively removing domain-specific factors, yielding only marginal gains.
In contrast, our text-anchored compression explicitly filters spurious variations, sustaining positive improvements regardless of backbone strength.

% Furthermore, we include DINOv2 as a self-supervised backbone to mitigate potential overlap between supervised ImageNet pretraining and DG benchmark domains, which could otherwise confound evaluation~\cite{Yu2023RethinkingTE}.
% The clear margin of our method over its LP baseline—relative to other methods—indicates that the improvements are not attributable to inherited exposure, but instead stem from the intended domain-invariance mechanism.
% Recent work~\cite{Yu2023RethinkingTE} notes that common DG benchmarks may be affected by data leakage, as supervised backbones could have encountered similar domains during large-scale pretraining. To partially control for this effect, we additionally evaluate on DINOv2, a self-supervised backbone less likely to exhibit such leakage. Our method maintains a clear margin over the LP baseline in this setting, suggesting that the observed improvements are not solely attributable to inherited exposure.

Recent work~\cite{Yu2023RethinkingTE} notes that some DG benchmarks may be affected by data leakage, as supervised backbones could have encountered similar domains during large-scale pretraining. To partially control for this effect, we additionally evaluate on DINOv2, a self-supervised backbone less likely to exhibit such leakage. While most methods remain close to the LP baseline in this setting, our method maintains a clear margin, suggesting that the gains are not solely attributable to inherited exposure, but instead stem from the intended domain-invariance mechanism.

\begin{figure}[t]
  \centering
  % \includegraphics[width=\linewidth]{figures/flow_n_rho.pdf}
  % \caption{(a) Information processing flow analysis on PACS (House class) (b) Self-awareness comparison}
  \includegraphics[width=0.7\linewidth]{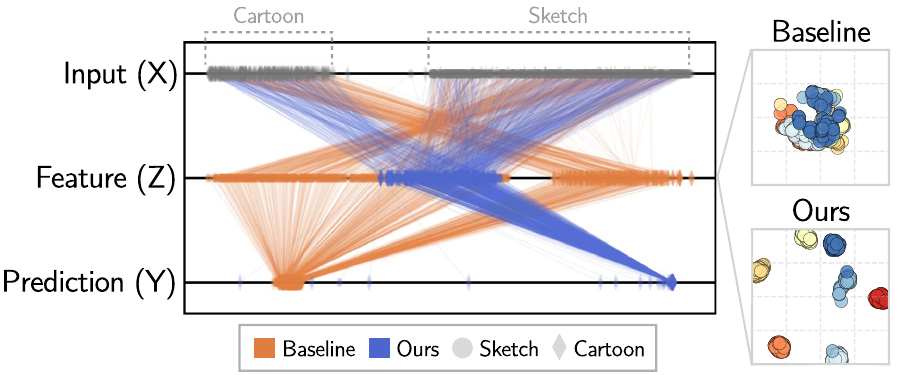}
  \caption{\textbf{Information flow ($X\rightarrow Z \rightarrow Y$) visualization using t-SNE on PACS.} 
  % left: cross-domain alignment for the \textit{house} class (ours vs. RISE). right: single-domain class-wise clusters.
  \textit{Left}: For the \textit{house} class, our method (blue) collapses inputs from different domains into a single cluster, while RISE (orange) retains domain-dependent separation, showing reduced cross-domain variation.
  \textit{Right}: Within a single domain, our embeddings form more compact, well-separated class clusters, showing sharper class margins.
  }
  \label{fig:flow}
\end{figure}

\subsection{Further Analysis}
\label{sec:exp:analysis}

\vspace{0.1cm} \noindent
\textbf{Ablation on Loss Components.}
\cref{tab:loss_main} ablates the three terms across three backbones on OfficeHome, PACS, and DomainNet.
Adding $\mathcal{L}_\text{comp}$ to $\mathcal{L}_\text{sem}$ drives the largest gains, while $\mathcal{L}_\text{align}$ provides a consistent further improvement; their combination performs the best across all backbones and datasets, confirming that compression and alignment are complementary.
App.~\ref{sec:app:loss_ablation} extends this ablation to other backbones.

\begin{table}[t]
\centering
\footnotesize
\caption{Ablation of loss components}
\label{tab:loss_main}
\setlength{\tabcolsep}{3.5pt}
\begin{tabular}{ccc|ccc|ccc|ccc}
\toprule
\multirow{2}{*}{$\mathcal{L}_{\text{sem}}$} &
\multirow{2}{*}{$\mathcal{L}_{\text{align}}$} &
\multirow{2}{*}{$\mathcal{L}_{\text{comp}}$} &
\multicolumn{3}{c|}{OfficeHome} & \multicolumn{3}{c|}{PACS} & \multicolumn{3}{c}{DomainNet} \\
& & & RN50 & ViT & CLIP & RN50 & ViT & CLIP & RN50 & ViT & CLIP \\
\midrule
\checkmark &            &            & 73.4 & 82.4 & 85.2 & 92.6 & 92.3 & 96.8 & 35.5 & 52.8 & 60.3 \\
\checkmark & \checkmark &            & 76.0 & 81.9 & 85.9 & 94.2 & 93.6 & 98.2 & 37.3 & 67.9 & 65.9 \\
\checkmark &            & \checkmark & 78.9 & 85.1 & 90.8 & 95.8 & 94.0 & 98.1 & 57.6 & 68.7 & 75.2 \\
\checkmark & \checkmark & \checkmark & \textbf{79.0} & \textbf{86.4} & \textbf{93.2} & \textbf{96.9} & \textbf{94.1} & \textbf{98.5} & \textbf{58.3} & \textbf{68.8} & \textbf{75.8} \\
\bottomrule
\end{tabular}
\end{table}

%\textbf{Text-Anchors Bottleneck Filters Spurious Domain Styles.}
\vspace{0.1cm} \noindent
\textbf{Does the bottleneck indeed suppress domain-specific information?}
To verify this,
%that the proposed bottleneck suppresses domain-specific cues,
we visualize the information flow for a specific class (\emph{house}) using t-SNE in~\cref{fig:flow}.
Across the cartoon and sketch domains in PACS, the inputs (top) are clearly domain-separated. 
Our features (blue in the middle, $Z$) concentrate into a unified cluster successfully, discarding domain information.
On the other hand, the baseline (RISE) features (orange in the middle, $Z$) largely retain the domain-dependent separation. 
Within each single domain, our embedding space (right) forms more compact class clusters, indicating that the text-anchored bottleneck removes cross-domain variation, enhancing class separability with larger margins.

\begin{figure}[t]
  \centering
  \includegraphics[width=\linewidth]{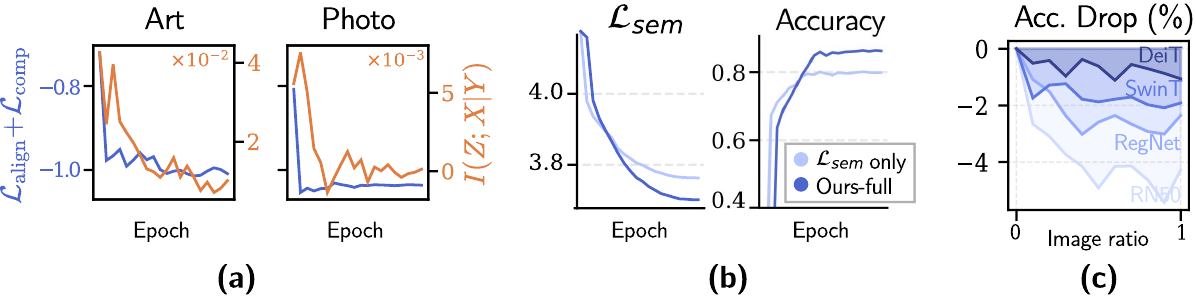}
  \caption{(a) $I(Z;X|Y)$ decreases with the regularizers $\mathcal{L}_\text{align}, \mathcal{L}_\text{comp}$ during training. (b) Training with the regularizers achieves lower semantic loss and higher accuracy. (c) Accuracy decreases as the image guidance ratio increases.}
  \label{fig:loss}
  % \vspace{-0.4cm}
\end{figure}

To examine the underlying mechanism, we track the CEB term $I(Z;X|Y)$ during training. 
\cref{fig:loss}(a) estimates this quantity using MINE~\cite{pmlr-v80-belghazi18a}. 
% In practice, $I(Z;X|Y)$ is approximated by $\mathcal{L}_{\text{align}}$ and $\mathcal{L}_{\text{comp}}$ in \cref{eq:loss_full}, and decreases together with them during training, indicating that the objective suppresses dependence on input-specific variations, as intended.
In practice, $I(Z;X|Y)$ is approximated by the alignment and compression terms ($\mathcal{L}_{\text{align}}$ and $\mathcal{L}_{\text{comp}}$ in \cref{eq:loss_full}), which decrease together during training, indicating that the objective suppresses dependence on input-specific variations.
This mechanism also facilitates semantic learning. 
\cref{fig:loss}(b) compares the full model with a variant trained only with $\mathcal{L}_{\text{sem}}$.
By filtering domain-specific variations, the model allows $\mathcal{L}_{\text{sem}}$ to converge to a lower value, resulting in higher accuracy.

% To examine the underlying mechanism \joonseok{of the bottleneck, we track its} training dynamics in \cref{fig:loss}.
% In (a), the estimated \eunyi{$I(Z;X|Y)$, using MINE~\cite{pmlr-v80-belghazi18a}, decreases alongside the regularizer losses ($\mathcal{L}_\text{align}$ and $\mathcal{L}_\text{comp}$) indicating that the derived objective effectively controls feature dependence on input variations.}
% \cref{fig:loss}(b) further compares \joonseok{the loss and achieved accuracy of our full model and a simpler one trained only with} %training with only
% the semantic objective $\mathcal{L}_\text{sem}$ (no compression).
% %against the full objective (Ours-full).
% By suppressing domain variations, the bottleneck allows $\mathcal{L}_\text{sem}$ to converge to a lower \joonseok{loss}, ultimately yielding higher accuracy.

\vspace{0.1cm} \noindent
%\textbf{Visual Guidance Reintroduces Domain Entanglement.}
\textbf{Does visual guidance reintroduce domain entanglement?}
In~\cref{fig:loss}(c), we test this by injecting CLIP image supervision into our objective in~\cref{eq:loss_full} with varied guidance ratios.
% We next examine the effect of reintroducing visual guidance to our method.
% In~\cref{fig:loss}(c), we inject CLIP image supervision into our objective in \cref{eq:loss_full} by varying the guidance ratio.
As the ratio increases, the accuracy consistently drops, with CNNs (RegNet, RN50) degrading more than Transformers (DeiT, SwinT), likely because such guidance propagates domain-specific statistics that CNNs are more sensitive to.
This indicates that aligning with expressive visual features disrupts the text-driven abstraction. 
Interestingly, this trend contrasts with prior KD approaches~\cite{Huang2023ASS,Addepalli_2024_CVPR}, where a small amount of textual guidance often improves performance, suggesting different roles of text under the two settings.

\begin{table*}[t]
\centering
\caption{Analysis of anchor design: (a) what to use and (b) how to integrate it}
\label{tab:proto}
\vspace{-0.4cm}
\begin{subtable}[t]{0.7\textwidth}
\centering
\caption{Effect of anchor encoder and template}
\vspace{-0.2cm}
\scriptsize
\setlength{\tabcolsep}{6pt}
\begin{tabular}{c l c}
\toprule
\textbf{Encoder} & \textbf{Template} & \textbf{Acc.} \\
\midrule
\multirow{2}{*}{CLIP (VL)} & `a photo of [cls]' & 79.0 \\
                      & ImageNet prompts~\cite{pmlr-v139-radford21a} & \textbf{77.9} \\
\midrule
\multirow{2}{*}{MiniLM (L)} & `a photo of [cls]' & 79.1 \\
                     & ImageNet prompts~\cite{pmlr-v139-radford21a} & 77.7 \\
\midrule
\multirow{2}{*}{MPNet (L)} & `a photo of [cls]' & 78.7 \\
                     & ImageNet prompts~\cite{pmlr-v139-radford21a} & 78.9 \\
\midrule
random & - & 76.6 \\
\bottomrule
\end{tabular}
\end{subtable}

\vspace{0.4cm}

\begin{subtable}[t]{0.7\textwidth}
\centering
\caption{Effect of anchor source and objective}
\vspace{-0.2cm}
\scriptsize
\setlength{\tabcolsep}{6pt}
\begin{tabular}{l cc cc cc}
\toprule
 & \multicolumn{2}{c}{\textbf{Contrastive}} & \multicolumn{2}{c}{\textbf{L2}} & \multicolumn{2}{c}{\textbf{Ours}} \\
\cmidrule(lr){2-3}\cmidrule(lr){4-5}\cmidrule(lr){6-7}
\textbf{Fixed?} & \xmark & \cmark & \xmark & \cmark & \xmark & \cmark \\
\midrule
\rowcolor{gray!10} \multicolumn{7}{l}{\textit{OfficeHome}} \\
random & 71.4 & 73.4 & 13.4 & 11.9 & 75.2 & 76.6 \\
CLIP   & 74.6 & 73.4 & 68.3 & 68.4 & 77.2 & \textbf{79.0} \\
\midrule
\rowcolor{gray!10} \multicolumn{7}{l}{\textit{PACS}} \\
random & 93.7 & 94.8 & 92.3 & 93.3 & 83.4 & 95.5 \\
CLIP   & 94.6 & 94.9 & 91.7 & 88.0 & 95.2 & \textbf{96.9} \\
\bottomrule
\end{tabular}
\end{subtable}

\end{table*}

\vspace{0.1cm} \noindent
\textbf{Why do text embeddings serve as effective anchors?}
% \textbf{Is our method agnostic to any language model?}
%\textbf{Fixed Anchors Provide Robust Invariance Priors.}
% Since our bottleneck relies on textual anchors, we first examine whether our framework depends on a specific CLIP property or reflects a more general characteristic of language spaces. 
% \cref{tab:proto}(a) compares anchors derived from different language encoders (MiniLM, MPNet) and prompt templates on OfficeHome. 
% All language encoders perform comparably well, and even random anchors achieve competitive performance. 
% This indicates that our framework is largely model-agnostic, leveraging the structural stability of anchor spaces rather than relying on CLIP-specific semantics.
% Further insights into how these language models capture class relationships are available in App.~\ref{sec:app:class_similarity}, where we compare the class similarity rankings produced by CLIP and MiniLM.
Since our bottleneck relies on textual anchors, we examine whether it depends on specific CLIP properties or reflects a more general characteristic of language spaces. 
\cref{tab:proto}(a) compares anchors from different language encoders (MiniLM~\cite{10.5555/3495724.3496209}, MPNet~\cite{10.5555/3495724.3497138}) and prompt templates on OfficeHome. 
All variants perform similarly, indicating that the framework is largely agnostic to particular language model and not tied to CLIP-specific representations 
% (see App.~\ref{sec:app:class_similarity} for class-similarity analysis).
(see App.~\ref{sec:app:rich_prompts} for richer prompt variants explored; App.~\ref{sec:app:class_similarity} for class-similarity analysis).
 
Even random anchors achieve competitive performance, surpassing the previous state-of-the-art (76.6 \textit{vs}.\ 74.4), suggesting that stable external anchors are the key to the framework’s effectiveness,
while pretrained text embeddings further benefit from their learned semantic structure.

Crucially, this semantic advantage grows as class discrimination becomes harder; on DomainNet (345 classes), the gap between CLIP and random anchors widens sharply (58.3 \textit{vs}. 36.4).
It is most evident in open-set generalization, where unseen test classes make random anchors inapplicable, yet text anchors still generalize well (78.8 \textit{vs}. CLIPood 78.2; 
\cref{tab:open_set}).
% App.~\ref{sec:app:os})

\begin{table}[t]
\footnotesize
\centering
\caption{Comparison on Open-Set DG.}
\vspace{-0.2cm}
\label{tab:open_set}
\scriptsize
\begin{tabular}{l cc}
\toprule
        & Known classes & Unseen classes \\
\midrule
CLIP    & 86.1 & 77.6 \\
CLIPood & 89.4 & 78.2 \\
Ours & \textbf{90.6} & \textbf{78.8} \\
\bottomrule
\end{tabular}
\end{table}
% To further understand how these anchors should be utilized, we analyze prototype design choices in \cref{tab:proto}(b). 
% On PACS, we compare CLIP and random anchors under different alignment strategies—including DPE~\cite{to2025diverse}, a prototype-ensemble classifier for robustness—and vary the learnability of the prototypes during training. 
% Allowing the prototypes to be updated significantly degrades performance, suggesting contamination from domain-specific training statistics. 
% Moreover, simple alignment objectives (\emph{e.g.}, L2) fail to capture relational semantics even with CLIP anchors. 
% These results indicate that domain-invariant learning benefits from fixed anchors combined with structured alignment, as incorporated in our objective.
% Finally, extensive ablations of the loss terms in~\cref{eq:loss_full} (App.~\ref{sec:app:loss_ablation}) and sensitivity analyses of $\beta_1, \beta_2$ (App.~\ref{sec:app:hp_sensitivity}) confirm the necessity of each component in our bottleneck formulation.

\vspace{0.1cm} \noindent
\textbf{How should anchors be integrated into training?}
\cref{tab:proto}(b)
% compares prototype strategies on PACS, including DPE~\cite{to2025diverse}, a method that aggregates multiple prototype classifiers for improved robustness.
varies three factors on OfficeHome and PACS: the anchor source (random \textit{vs}. CLIP), whether anchors are kept fixed (\cmark), and the alignment objective.
% Allowing prototypes trainable (\checkmark) degrades performance, as anchors absorb domain-specific statistics and drift from domain-invariant references.
Freezing inconsistently helps for contrastive and L2 while consistently under ours, as learnable anchors otherwise absorb domain-specific statistics and drift from invariant references.
% We also compare several alignment objectives.
% While simple losses such as contrastive or L2 remain suboptimal, our CEB-based objective better preserves semantic structure via compression and directional constraints.
Likewise, our CEB-based objective is the only one that turns semantic CLIP anchors into reliable gains, via compression and directional constraints.
Overall, domain invariance emerges only when stable external anchors are paired with an objective that explicitly enforces such invariance.

\section{Conclusion}
\label{sec:6-conclusion}

% We challenge the common assumption in DG that increasing representational expressiveness improves robustness. Instead, we show that expressive visual features propagate spurious domain cues. We propose a Text-Anchored Information Bottleneck that enforces invariance through stable textual anchors, eliminating conventional visual teachers.
% More fundamentally, our results suggest that invariance is governed not by representation capacity but by the nature of supervision. Rather than searching for shared correlations across domains, we ground guidance in external, domain-stable anchors to explicitly suppress domain-specific factors.
% However, fixed anchors become limiting when correct identification requires contextual cues. Moreover, like existing benchmarks, our framework does not fully disentangle performance gains from the broader issue of pretrained data leakage. Future work will explore adaptive anchor structures and alternative signals to dynamically disentangle core semantics from domain-specific contexts.
Expressive visual features can propagate spurious cues, challenging the common belief that greater capacity ensures robustness.
We introduce a Text-Anchored Information Bottleneck that explicitly grounds supervision in external anchors to directly address the source of domain bias.
Rather than distilling invariance from domain-shifting visual data, we utilize stable external signals to prove that the structural source of supervision is more critical than model capacity.
While fixed anchors lack contextual cues and benchmark data leakage~\cite{Yu2023RethinkingTE} currently obscures true generalization gains, it will be an interesting future direction to explore adaptive anchors and alternative signals to resolve these limitations and better isolate core semantics.

% \etal, \eg
% \begin{itemize}
% \item I have removed all \verb| \vspace| and \verb|\hspace|  commands from my paper.
% \item I have not used \verb|\cite| command in the abstract.
% \item I have entered a correct \verb|\titlerunning{}| command and selected a meaningful short name for the paper.
% \item I have used the same name spelling in all my papers accepted to ECCV and ECCV Workshops.
% \item I have added acknowledgments without a section number, e.g. using the \verb|\section*{}| command.
% \item Excluding references and acknowledgments, my paper is no longer than 14 pages.
% \item I have not decreased the font size of any part of the paper (except tables) to fit into 14 pages, I understand Springer editors will remove such commands.
% \end{itemize}

% \clearpage\mbox{}Page \thepage\ of the manuscript.
% \clearpage\mbox{}Page \thepage\ of the manuscript.
% \clearpage\mbox{}Page \thepage\ of the manuscript.
% \clearpage\mbox{}Page \thepage\ of the manuscript.
% \clearpage\mbox{}Page \thepage\ of the manuscript. This is the last page.
% Now we have reached the maximum length of an ECCV \ECCVyear{} submission (excluding references and acknowledgements).
% References should start immediately after the main text, but can continue past p.\ 14 if needed. 
% \clearpage  % TODO FINAL: This \clearpage needs to be removed from both review and camera-ready versions.

% \section*{Acknowledgements}
% Please insert your acknowledgments here.


\section*{Acknowledgments}

This work was supported by Samsung Electronics, Youlchon Foundation, National Research Foundation of Korea (NRF) grants (RS-2021-NR05515, RS-2024-00336576, RS-2023-0022663), and the Institute for Information \& Communication Technology Planning \& Evaluation (IITP) grants (RS-2022-II220264, RS-2024-00353131) funded by the Korean government.

\bibliographystyle{splncs04}
\bibliography{main}

\clearpage
\appendix
\label{sec:7-appendix}

\crefname{section}{App.}{Apps.}
\Crefname{section}{App.}{Apps.}

\setcounter{page}{1}
\pagenumbering{roman}
\crefalias{section}{appendix}
\crefalias{subsection}{appendix}
\crefalias{subsubsection}{appendix}
\section*{Appendix}

\setcounter{table}{0}
\setcounter{figure}{0}
\renewcommand{\thetable}{\Roman{table}}
\renewcommand{\thefigure}{\Roman{figure}}

%\textbf{Appendix Overview.}
% Sec.~\ref{sec:app:caption} provides details on caption generation used in our text-embedding analysis.
% Sec.~\ref{sec:app:img_guidance} presents an information-theoretic analysis of image guidance in domain generalization.
% Sec.~\ref{sec:app:hp_sensitivity} examines hyperparameter robustness across datasets.
% Sec.~\ref{sec:app:exp_details} summarizes implementation details.
% Sec.~\ref{sec:app:backbones} provides additional experimental results on diverse backbones, and Sec.~\ref{sec:app:failure} provides failure analysis.
% Secs.~\ref{sec:app:class_similarity} and~\ref{sec:app:loss_ablation} provide further analysis of class similarity across text encoders and loss ablations across backbones.

\section{Details on Caption Generation}
\label{sec:app:caption}

\begin{figure}
\centering
\includegraphics[width=\columnwidth]{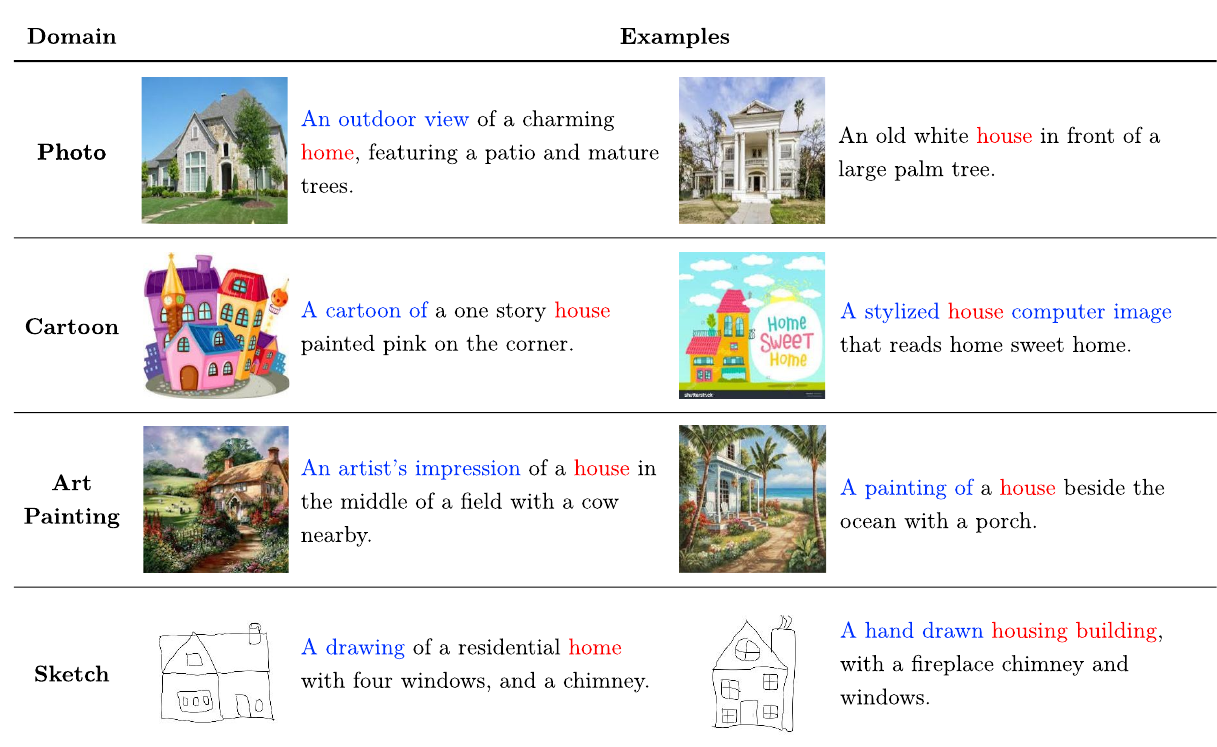}
\caption{\textbf{Example of generated captions in \textit{house} class, across four PACS domains}. \textcolor{red}{Red} highlights the semantic class (\emph{e.g.}, \textit{house, home}), \textcolor{blue}{blue} indicates domain-specific cues, and remaining denotes supplementary details unrelated to class semantics.}
\label{fig:captions}
\end{figure}

To better understand the structure of CLIP text embeddings, we visualize image-conditioned captions generated using \texttt{LLaVA-v1.5-7B}\footnote{H. Liu et al., Improved Baselines with Visual Instruction Tuning, CVPR 2024.} in \cref{fig:captions}. 

Each image from the PACS dataset is fed to the captioner with the following prompt:

\begin{quote}
``\textbf{USER:} \texttt{<image>} Generate a short and clear caption for the image. \textbf{ASSISTANT: }''.
\end{quote}
The generated captions capture a mix of high-level semantic concepts (\emph{e.g.}, \emph{house}, \emph{home}) and domain-specific visual styles (\textit{e.g.}, \emph{a painting of}, \emph{a cartoon of}), along with peripheral details present in the images.
As shown in~\cref{fig:captions}, for example, the surrounding descriptions somewhat vary in style and content, but the core semantics associated with the class label (\textit{e.g.}, \emph{house}) is consistently preserved in the textual embeddings across all domains.

\section{Information-Theoretic Analysis}
\label{sec:app:img_guidance}
We formalize the intuition in Sec.~\ref{sec:3-motivation} that image-dependent guidance can increase reliance on spurious domain information, thereby reducing the predictive information available for unseen-domain generalization.

\begin{figure}[t]
    \centering
    \begin{tikzpicture}[
        node distance=1.5cm,
        >=Stealth,
        every node/.style={draw, circle, inner sep=2pt, minimum size=3em},
        edge/.style={->, thick}
    ]

    \node (X_sp) at (-2, 1.5) {$X_{\text{sp}}$};
    \node (X_inv) at (0, 1.5) {$X_{\text{inv}}$};
    \node (Y) at (2, 1.5) {$Y$};
    \node (X) at (-1, 0) {$X$};
    \node (S) at (1, 0) {$S$};
    
    \draw[edge] (X) -- (S);
    \draw[edge] (X_inv) -- (X);
    \draw[edge] (X_inv) -- (Y);
    \draw[edge] (X_sp) -- (X);

    \end{tikzpicture}
    \caption{\textbf{Causal graph of the domain generalization setting}. Images are generated from invariant semantic factors $X_\text{inv}$ and domain-specific factors $X_\text{sp}$. Only $X_\text{inv}$ causally determines the label $Y$, while $X_\text{sp}$ introduces spurious correlations.}
    \label{fig:causal}
\end{figure}
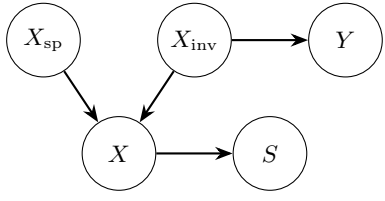

\vspace{0.1cm}\noindent
\textbf{Notation.}
An input image $X$ can be decomposed into domain-invariant and domain-specific components: $X = (X_{\text{inv}}, X_{\text{sp}})$.
We denote the class label by $Y$ and the student representation by $S = \phi(X)$.
The text teacher provides guidance $G_{\text{t}}=g_{\text{t}}(Y)$, while the image teacher provides guidance $G_{\text{i}} = g_{\text{i}}(X)$.
Their mixture is denoted by $G_{\alpha} = \alpha G_{\text{i}}+(1-\alpha)G_{\text{t}}$, with a weight $\alpha \in [0,1]$.

\vspace{0.1cm}\noindent
\textbf{Assumptions.}
We assume the following:
\begin{itemize}
    \item (A1) $X$ is generated from $X_\text{inv}$ and $X_\text{sp}$, the label $Y$ is determined by $X_\text{inv}$, and $S$ is a function of $X$. Consequently, $Y \perp S \mid X_{\text{inv}}$ and $X_{\text{inv}} \perp X_{\text{sp}}$ (see \cref{fig:causal}).
    \item (A2) The text guidance is independent of the spurious component, whereas image guidance is not, i.e., $I(G_{\text{t}};X_{\text{sp}})=0$ and $I(G_{\text{i}};X_{\text{sp}})>0$. This reflects our setting, where $G_{\text{t}}$ is conditioned on $Y$, whereas $G_{\text{i}}$ depends on $X$.
    
    \item (A3) As $\alpha$ increases, spurious information in the mixed guidance is non-decreasing and is partially transferred to the learned representation. Formally, we assume that $I(G_\alpha;X_\text{sp})$ is non-decreasing in $\alpha$, and that there exists $c\in(0,1]$ such that $I(S;X_\text{sp}) \approx c\, I(G_\alpha;X_\text{sp})$.
\end{itemize}

\vspace{0.1cm} \noindent
\textbf{Proposition 1.} Under (A2)--(A3), $I(S; X_{\mathrm{sp}})$ is non-decreasing in $\alpha$.

% $\frac{\partial}{\partial \alpha} I(S; X_{\mathrm{sp}}) \ge 0$.
\begin{proof}
By (A2), a larger $\alpha$ increases $I(G_\alpha; X_{\mathrm{sp}})$.
Then, by (A3), this implies that $I(S; X_{\mathrm{sp}})$ also increases.
\end{proof}

\vspace{0.1cm} \noindent
\textbf{Proposition 2.} Let $C$ denote the model capacity of $\phi$. Then the information encoded in the representation is bounded as $I(S;X) =(S;X_{\mathrm{inv}})+I(S;X_{\mathrm{sp}}\mid X_{\mathrm{inv}})\le C$.
\begin{proof}
By the chain rule and the decomposition $X = (X_{\mathrm{inv}}, X_{\mathrm{sp}})$, we have:
\begin{equation}
I(S;X) = I(S;X_{\mathrm{inv}}, X_{\mathrm{sp}}) = I(S;X_{\mathrm{inv}}) + I(S;X_{\mathrm{sp}} \mid X_{\mathrm{inv}}).
\end{equation}
Since $S = \phi(X)$ is deterministic, $H(S|X) = 0$ and $I(S;X) = H(S)$. Thus,
\begin{equation}
I(S;X) = H(S) \le C.
\end{equation}
\end{proof}

\vspace{0.1cm} \noindent
\textbf{Proposition 3.} Under $X_{\mathrm{inv}} \perp X_{\mathrm{sp}}$ (A1), 
$I(S;X_{\mathrm{sp}} \mid X_{\mathrm{inv}})\ge I(S;X_{\mathrm{sp}})$.
\begin{proof}
Equating two chain-rule expansions of $I(S;X_{\mathrm{sp}},X_{\mathrm{inv}})$,
\begin{align}
I(S;X_{\mathrm{inv}},X_{\mathrm{sp}})
&= I(S;X_{\mathrm{inv}}) + I(S;X_{\mathrm{sp}} \mid X_{\mathrm{inv}}) \\
&= I(S;X_{\mathrm{sp}}) + I(S;X_{\mathrm{inv}} \mid X_{\mathrm{sp}}),
\end{align}
yields
\begin{equation}
I(S;X_{\mathrm{sp}} \mid X_{\mathrm{inv}})
=
I(S;X_{\mathrm{sp}})
+ I(S;X_{\mathrm{inv}} \mid X_{\mathrm{sp}})
- I(S;X_{\mathrm{inv}}).
\label{eq:mu-s-sp}
\end{equation}
Similarly, equating two chain-rule expansions of $I(S;X_{\mathrm{sp}};X_{\mathrm{inv}})$,
\begin{align}
I(S;X_{\mathrm{sp}};X_{\mathrm{inv}})
&= I(S;X_{\mathrm{inv}})-I(S;X_{\mathrm{inv}} \mid X_{\mathrm{sp}}) \\
&= I(X_{\mathrm{inv}};X_{\mathrm{sp}})-I(X_{\mathrm{inv}};X_{\mathrm{sp}} \mid S),
\end{align}
gives
\begin{equation}
I(S;X_{\mathrm{inv}} \mid X_{\mathrm{sp}})
=
I(S;X_{\mathrm{inv}})
+ I(X_{\mathrm{inv}};X_{\mathrm{sp}} \mid S)
- I(X_{\mathrm{inv}};X_{\mathrm{sp}}).
\label{eq:mu-triple}
\end{equation}
Substituting \cref{eq:mu-triple} into \cref{eq:mu-s-sp} yields
\begin{equation}
I(S;X_{\mathrm{sp}} \mid X_{\mathrm{inv}})
=
I(S;X_{\mathrm{sp}})
+ I(X_{\mathrm{inv}};X_{\mathrm{sp}} \mid S)
- I(X_{\mathrm{inv}};X_{\mathrm{sp}}).
\end{equation}
Since $X_{\mathrm{inv}} \perp X_{\mathrm{sp}}$ by (A1), $I(X_{\mathrm{inv}};X_{\mathrm{sp}})=0$. Thus
\begin{equation}
I(S;X_{\mathrm{sp}} \mid X_{\mathrm{inv}})
=
I(S;X_{\mathrm{sp}})
+ I(X_{\mathrm{inv}};X_{\mathrm{sp}} \mid S)
\ge I(S;X_{\mathrm{sp}}),
\end{equation}
where the inequality follows from the non-negativity of mutual information.
\end{proof}

\vspace{0.1cm} \noindent
\textbf{Proposition 4.} $I(S;X_\text{inv})\ge I(S;Y)$.
\begin{proof}
By expanding the chain rule to $I(S;X_\text{inv}, Y)$:
\begin{align}
I(S;X_\text{inv}, Y)&=I(S;X_\text{inv})+{I(S;Y|X_\text{inv})}\\
&= I(S;Y)+I(S;X_\text{inv}|Y),
\end{align}
it follows that:
\begin{equation}
I(S;X_\text{inv}) = I(S;Y) +I(S;X_\text{inv}|Y)-I(S;Y|X_\text{inv}).
\end{equation}
By A1, $I(S;Y|X_\text{inv})= 0$. Thus, due to the non-negativity of mutual information, we obtain $I(S;X_\text{inv})\geq I(S;Y)$.
\end{proof}

\vspace{0.1cm} \noindent
\textbf{Proposition 5.} A decrease in $I(S;Y)$ increases the lower bound on the test error $\mathcal{E}_\text{test}$.
\begin{proof}
By Fano's inequality~\cite{fano1961transmission},
\begin{equation}
H(Y\mid S)=H(Y)-I(S;Y)
\le h(\mathcal{E}_{\text{test}})+\mathcal{E}_{\text{test}}\log(|\mathcal Y|-1),
\label{eq:prop4-1}
\end{equation}
where $h(\cdot)$ denotes the binary entropy and $|\mathcal{Y}|$ is the number of classes. Using $h(\mathcal{E}_{\text{test}})\le 1$, we obtain
\begin{equation}
\mathcal{E}_{\text{test}}
\ge
\frac{H(Y\mid S)-1}{\log |\mathcal Y|-1}.
\label{eq:prop4-2}
\end{equation}
Therefore, decreasing $I(S;Y)$ increases $H(Y\mid S)$ (\cref{eq:prop4-1}), and hence increases the lower bound on $\mathcal{E}_{\text{test}}$ (\cref{eq:prop4-2}).
\end{proof}

\vspace{0.1cm} \noindent
\textbf{Interpretation.}
Propositions 1--5 collectively delineate a trade-off mechanism between image guidance and model generalization. 
Specifically, P1 demonstrates that increasing the weight on image guidance causes the representation $S$ to absorb more spurious information $X_\text{sp}$.
Under finite capacity, this reduces the information budget available for $I(S;X_\text{inv})$ (P2--3).
Consequently, the label-predictive information in the representation is diminished (P4), which, via Fano’s inequality, leads to a higher lower bound on the test error (P5).
In summary,
\begin{equation}
\alpha \uparrow \;\Rightarrow\; I(S; X_\text{sp}) \uparrow \;\Rightarrow\; I(S; X_\text{inv}) \downarrow \;\Rightarrow\; I(S; Y) \downarrow \;\Rightarrow\; \mathrm{LB}(\mathcal{E}_\text{test}) \uparrow.
\end{equation}

\section{Hyperparameter Sensitivity}
\label{sec:app:hp_sensitivity}

\begin{figure}[t]
    \centering
    \includegraphics[width=0.45\columnwidth]{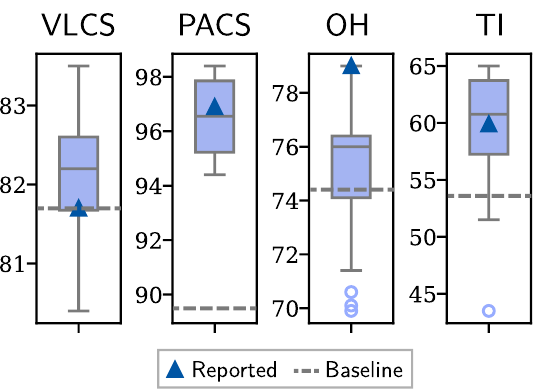}
    \caption{\textbf{Sensitivity to $(\beta_1,\beta_2)$ across datasets}. Boxplots show results over searched hyperparameter combinations; triangle: fixed setting, dashed line: best SOTA baseline \eunyi{(VLCS: RISE, PACS: RISE, OH: VL2V, TI: VL2V)}.}
    \label{fig:hp}
\end{figure}

In~\cref{fig:hp}, we evaluate the sensitivity of our framework to the hyperparameters $\beta_1$ and $\beta_2$, searched over \{0.0, 0.1, 0.5, 1.0\}.
Based on the best ResNet-50 validation accuracy on PACS, we select a fixed configuration of $\beta_1=0.1$ and $\beta_2=1.0$.

Using this single configuration (denoted by the triangle in~\cref{fig:hp}), our method remains competitive to, and often surpasses, the strongest SOTA baselines (dashed line).
Although dataset-specific tuning could yield marginal gains on certain benchmarks such as VLCS or TerraIncognita, we intentionally keep these hyperparameters fixed across all datasets and architectures.

This choice is motivated by recent observations that extensive hyperparameter tuning may introduce evaluation biases in domain generalization benchmarks~\cite{Yu2023RethinkingTE}.
While we do not modify the standard evaluation protocol, the results suggest strong performance under a shared hyperparameter configuration.

\section{Standard Deviation across Seeds}
\label{sec:app:std}
All results are averaged over three seeds; \cref{tab:std} reports per-method standard deviations on the CLIP-ViT-B/16 backbone.
\begin{table}[t]
\centering
\caption{Standard deviation with CLIP-ViT-B/16.}
\label{tab:std}
\begin{tabular}{l|ccccc}
\toprule
Method  & VLCS & PACS & OH & TI & DN \\
\midrule
MIRO    & 82.2$_{\pm0.3}$ & 95.6$_{\pm0.8}$ & 82.5$_{\pm0.1}$ & 54.3$_{\pm0.4}$ & 54.0$_{\pm0.3}$ \\
CLIPood & 85.0$_{\pm0.4}$ & 97.3$_{\pm0.1}$ & 87.0$_{\pm0.2}$ & 60.4$_{\pm0.7}$ & 63.5$_{\pm0.1}$ \\
VL2V    & 83.3$_{\pm0.4}$ & 96.7$_{\pm0.6}$ & 87.4$_{\pm0.3}$ & 58.5$_{\pm0.7}$ & 62.8$_{\pm0.1}$ \\
Ours & 89.0$_{\pm0.4}$ & 98.5$_{\pm0.1}$ & 93.2$_{\pm0.3}$ & 75.1$_{\pm0.6}$ & 75.8$_{\pm0.2}$ \\
\bottomrule
\end{tabular}
\end{table}

\section{More Experimental Details}
\label{sec:app:exp_details}

Following common practice, we train for 5K iterations with batch size 32 (64 for OfficeHome) for TerraIncognita, OfficeHome, PACS, and VLCS, and for 15K for DomainNet and NICO\textsuperscript{++} with a batch size of 64.
We use cosine annealing without warmup.
Learning rates are set to $10^{-4}$ for ResNet/EfficientNet, $10^{-6}$ for CLIP backbones, and $10^{-5}$ otherwise. 

Our mapper is implemented as a lightweight Transformer encoder that projects image features into the text embedding space, and its learning rate is fixed to $10^{-4}$ in all experiments.
To stabilize training on OfficeHome, we freeze the pretrained image backbone for the first 1K iterations and optimize only the mapper before full end-to-end training.
Our implementation is based on PyTorch 1.13 with CUDA 11.8, and all experiments are conducted on a single NVIDIA RTX A6000 GPU.
% Our method performs calculations on a per-class basis. To support this, our dataloader samples a fixed number of random class labels per iteration and draws an equal number of samples from each class. For example, with a default batch size of 32, we sample 4 random classes with 8 samples per class. For OfficeHome, where the batch size is 64, we sample 16 random classes with 4 samples each.

We report MIRO, CLIPood, RISE, and VL2V results using their official codebases, and extend them to previously unreported backbones.
For these unreported backbone settings, we tune only the learning rate around each method’s default values. Specifically, we search \{$5 \cdot 10^{-6}$, $10^{-5}$, $3 \cdot 10^{-5}$, $5 \cdot 10^{-5}$, $10^{-4}$\} for MIRO (default: \{$10^{-5}$, $3 \cdot 10^{-5}$\}),
\{$10^{-5}$, $5 \cdot 10^{-5}$, $10^{-4}$, $5 \cdot 10^{-4}$, $10^{-3}$\} for RISE (default: \{$10^{-3}$\}),
and \{$5 \cdot 10^{-6}$, $10^{-5}$, $5 \cdot 10^{-5}$, $10^{-4}$\} for both CLIPood (default: \{$5 \cdot 10^{-6}$, $10^{-5}$\}) and VL2V (default: \{$5 \cdot 10^{-5}$\}).
Model selection for these baselines follows the standard DomainBed protocol used in prior work, where hyperparameters are selected separately for each target-domain split. In contrast, our method uses a single universal hyperparameter setting across domains, datasets and architectures, as described in \cref{sec:app:hp_sensitivity}.

\section{Performance Across Diverse Backbones}
\label{sec:app:backbones}

\begin{table}[t]
\centering
\footnotesize
\setlength{\tabcolsep}{3pt}
\caption{\textbf{DG performance across diverse backbones and pretraining sources}. Parentheses indicate the pretraining source.}
\begin{tabular}{l | c c c c |c || c c c c |c}

\toprule
\textbf{Method}  & \textbf{VLCS} & \textbf{PACS} & \textbf{OH} & \textbf{TI} & \textbf{Avg}  & \textbf{VLCS} & \textbf{PACS} & \textbf{OH} & \textbf{TI} & \textbf{Avg} \\ 
\midrule
\rowcolor{gray!10} \multicolumn{6}{c}{\textit{EfficientNet (IN-1K)}} & \multicolumn{5}{c}{\textit{Swin Transformer (IN-21K)}}\\ 
% RegNetY Backbone
LP & 80.7 & 72.2 & 69.0 & 39.6 & 65.4 & 78.4 & 93.4 & \underline{86.6} & 55.5 & 78.5 \\ 
MIRO & 80.2 & 83.2 & 72.4 & 40.6 & 69.1 & 82.0 & 93.5 & 83.5 & \underline{57.7} & 79.2\\ 
CLIPood & 79.1 & \underline{91.2} & 71.1 & 43.8 & 68.7 & 76.9 & 86.2 & 80.9 & 40.4 & 71.1\\ 
RISE & \underline{80.9} & 88.0 & 68.4 & 45.1 & 70.6 & \underline{84.3} & \underline{94.2} & 82.8 & 53.4 & 78.7\\ 
VL2V &  80.5 & 86.8 & \underline{75.3} & \textbf{47.2} & \underline{72.5} & 82.4 & 93.1 & 84.8 & 57.0 & \underline{79.3} \\ 
\rowcolor{cyan!10}
Ours & \textbf{84.0} &  \textbf{92.3} &  \textbf{75.7} &  \underline{46.8} &  \textbf{74.7} & \textbf{88.8} &  \textbf{97.7} &  \textbf{92.9} &  \textbf{78.4} &  \textbf{89.4} \\ 
\midrule

% Effnet Backbone
\rowcolor{gray!10} \multicolumn{6}{c}{\textit{RegNetY-16GF (IN-12K)}} & \multicolumn{5}{c}{\textit{DeiT (IN-1K)}}\\ 
LP  &  78.1 & 86.2 & 68.4 & 46.3 & 69.7 & 79.9 & 89.4 & 77.6 & 49.5 & 74.1 \\ 
MIRO &  79.0 & 85.4 & 70.5 & 50.4 & 71.3 & 80.1 & 83.7 & 78.1 & 49.9 & 72.9 \\ 
CLIPood &  81.7 & \underline{94.1} & 81.1 & \underline{57.5} & 78.6 & 81.0 & 90.6 & 77.3 & 49.8 & 74.7\\ 
RISE &  82.3 & 89.1 & 74.9 & 35.1 & 70.4 & \underline{83.5} & 90.6 & 74.8 & 46.9 & 74.0 \\ 
VL2V &  \underline{83.0} & 93.1 & \underline{83.6} & 55.3 & \underline{78.8} & 81.5 & \underline{90.8} & \underline{80.5} & \underline{53.4} & \underline{76.6} \\ 
\rowcolor{cyan!10}
Ours &  \textbf{85.4} &  \textbf{99.4} &  \textbf{87.2} &  \textbf{73.5} &  \textbf{86.4} &  \textbf{88.1} &  \textbf{95.8} &  \textbf{88.7} &  \textbf{71.5} &  \textbf{86.0} \\ 
\midrule

% RN-18 Backbone
\rowcolor{gray!10} \multicolumn{6}{c}{\textit{RegNetY-16GF (IG-3B)}} & \multicolumn{5}{c}{\textit{DINOv2 (LVD-142M)}}\\ 
LP & 81.0 & 92.4 & 81.3 & 55.2 & 77.5 & 82.6 & 95.8 & 84.5 & 57.4 & 80.1 \\ 
MIRO & 79.9 & 97.4 & 80.4 & 58.9 & 79.2  & 82.6 & 95.3 & \underline{85.1} & 60.4 & 80.9 \\ 
CLIPood & 81.6 & \underline{97.8} & 83.3 & \underline{62.5} & \underline{81.3} & 82.4 & \underline{96.8} & 81.6 & 58.1 & 79.7 \\ 
RISE & \underline{82.8} & 95.5 & 81.5 & 60.2 & 80.0 & 81.4 & 88.3 & 69.0 & 40.2 & 69.7 \\ 
VL2V & 82.7 & 96.7 & \underline{84.0} & 61.1 & 81.1 & \underline{83.6} & 95.1 & \underline{85.1} & \underline{61.6} & \underline{81.3}\\ 
\rowcolor{cyan!10}
Ours & \textbf{88.5} & \textbf{99.7} & \textbf{88.4} & \textbf{76.7} & \textbf{88.3} &  \textbf{87.9} &  \textbf{97.7} &  \textbf{90.2} &  \textbf{63.6} &  \textbf{84.9}\\
\bottomrule
\end{tabular}
\label{tab:backbone-all}
\end{table}

\cref{tab:backbone-all} provides the detailed results corresponding to \cref{fig:backbones} in the main text, covering diverse CNN and Transformer backbones with different pretraining sources.

\section{Failure Analysis: Trade-off with Visual Context}
\label{sec:app:failure}

\begin{figure}
\centering
\includegraphics[width=\columnwidth]{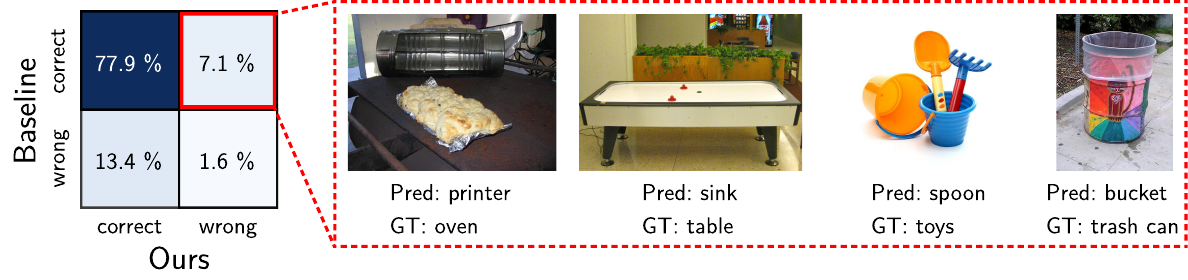}
\caption{\textbf{Comparison between our method and VL2V on OfficeHome}. Examples are drawn from the 7.1\% subset where VL2V predicts correctly and ours fails, often involving visually ambiguous objects for which broader image cues aid disambiguation.}
\label{fig:failures}
\end{figure}

We compare the predictions of VL2V and of our method on OfficeHome in~\cref{fig:failures}.
Our method fails on 7.1\% of samples where VL2V is correct, while it succeeds on 13.4\% of samples where VL2V fails, suggesting that image guidance is overall more harmful than beneficial under domain shift.

We further inspect the 7.1\% failure subset in~\cref{fig:failures}. 
These cases often involve visually ambiguous objects whose local appearance alone is not sufficiently distinctive. 
For example, our method predicts \textit{printer} for an \textit{oven}, \textit{spoon} for \textit{toys}, and \textit{bucket} for \textit{trash can}. 
Broader image cues can then help disambiguate the correct class. 
These examples highlight a trade-off of our approach: reducing non-essential visual variation improves overall robustness, but can hurt recognition when additional visual cues are needed for disambiguation.

\section{Loss Ablation Across Backbones}
\label{sec:app:loss_ablation}
\begin{table}
\centering
\footnotesize
\caption{Ablation of loss components on OfficeHome across diverse backbones.}
\label{tab:loss_ablation}
\begin{tabular}{c c c|c c c c c c c c c}
\toprule
$\mathcal{L}_{\text{sem}}$ & $\mathcal{L}_{\text{align}}$ & $\mathcal{L}_{\text{comp}}$ & RN-50 & ViT & CLIP & EffNet & Reg-IN & Reg-IG & DeiT & SwinT & DINO \\
\midrule
$\checkmark$ & & & 73.4 & 82.4 & 85.2 & 71.3 & 82.5 & 84.9 & 80.7 & 87.9 & 85.0 \\
$\checkmark$ & $\checkmark$ & & 76.0 & 81.9 & 85.9 & 75.0 & 82.7 & 83.6 & 81.2 & 86.4 & 83.7 \\
$\checkmark$ & & $\checkmark$ &  78.9 & 85.1 & 90.8 & \textbf{76.7} & 83.8 & 87.0 & 85.7 & 90.0 & 85.4 \\
$\checkmark$ & $\checkmark$ & $\checkmark$ & \textbf{79.0} & \textbf{86.4} & \textbf{93.2} & 75.7 & \textbf{87.2} & \textbf{88.4} & \textbf{88.7} & \textbf{92.9} & \textbf{90.2} \\
\bottomrule
\end{tabular}
\end{table}
To further validate the effectiveness of our loss design across backbones,~\cref{tab:loss_ablation} extends the ablation study to diverse architectures.
Starting from $\mathcal{L}_{\text{sem}}$, while adding $\mathcal{L}_{\text{comp}}$ generally is more impactful than adding $\mathcal{L}_{\text{align}}$ alone, their combination enforces directional consistency and reduces intra-class variance, facilitating convergence to higher accuracy across all backbones.

\section{Exploration on Prompt Types}
\label{sec:app:rich_prompts}

We evaluate richer, AI-generated prompts on four benchmarks. 
We consider two richer description styles, one based on lexical and hierarchical semantics and another on class-specific shape and function, against the simple class template.
As shown in Tab.~\ref{tab:app:rich_prompts}, increasing prompt richness yields no consistent gain over the simple template, indicating that class-level prompts already suffice for separability.

\begin{table}[t]
\centering
\caption{Exploration on prompt types.}
\label{tab:app:rich_prompts}
\begin{tabular}{l|cccc|c}
\toprule
Prompt & \textsc{vlcs} & \textsc{pacs} & \textsc{oh} & \textsc{ti} & Avg \\
\midrule
AI caption (lexical/hierarchical) & 83.3 & 94.0 & 77.3 & 54.9 & 77.4 \\
AI caption (shape/function)       & 82.3 & 95.1 & 79.4 & 59.7 & 79.1 \\
\rowcolor{gray!15} `a photo of a [cls]' & 81.7 & 96.9 & 79.0 & 59.9 & \textbf{79.4} \\
\bottomrule
\end{tabular}
\end{table}

\begin{table}
\centering
\caption{Top-5 most similar classes for selected OfficeHome queries under CLIP and MiniLM text embeddings.}
\label{tab:clip_sbert_comparison}
\footnotesize
\begin{tabular}{p{2cm} p{10cm}}
\toprule
\textbf{Query} & \textbf{Top-5 Similar Classes (CLIP / MiniLM)} \\
\midrule

\multirow{2}{*}{\textbf{candles}} 
& \textbf{CLIP:} knives, bottle, flowers, shelf, toys \\
& \textbf{MiniLM:} lamp shade, desk lamp, flowers, pencil, bed \\
\addlinespace

\multirow{2}{*}{\textbf{clipboards}} 
& \textbf{CLIP:} folder, notebook, laptop, shelf, calendar \\
& \textbf{MiniLM:} paper clip, scissors, post-it notes, pen, pencil \\
\addlinespace

\multirow{2}{*}{\textbf{couch}} 
& \textbf{CLIP:} bed, TV, chair, keyboard, shelf \\
& \textbf{MiniLM:} chair, bed, curtains, desk lamp, TV \\
\addlinespace

\multirow{2}{*}{\textbf{folder}} 
& \textbf{CLIP:} computer, calendar, notebook, laptop, printer \\
& \textbf{MiniLM:} file cabinet, trash can, paper clip, notebook, shelf \\
\addlinespace

\multirow{2}{*}{\textbf{fork}} 
& \textbf{CLIP:} pen, knives, toothbrush, scissors, pencil \\
& \textbf{MiniLM:} spoon, knives, mug, scissors, bike \\
\addlinespace

\multirow{2}{*}{\textbf{hammer}} 
& \textbf{CLIP:} drill, radio, pen, speaker, computer \\
& \textbf{MiniLM:} drill, screwdriver, knives, eraser, mug \\
\addlinespace

\multirow{2}{*}{\textbf{helmet}} 
& \textbf{CLIP:} kettle, bucket, backpack, mug, hammer \\
& \textbf{MiniLM:} backpack, glasses, bike, hammer, webcam \\

\bottomrule
\end{tabular}

\end{table}

\section{Class Similarity Comparisons Across Text Encoders}
\label{sec:app:class_similarity}

In \cref{tab:clip_sbert_comparison}, we report the top-5 nearest classes for several OfficeHome query labels using CLIP and MiniLM text embeddings. 
The two encoders capture similarity in different ways. 
CLIP often favors visual or shape-based resemblance; for example, for \textit{helmet}, it retrieves \textit{kettle} and \textit{bucket}, and for \textit{fork}, it retrieves \textit{pen} and \textit{knives}. 
In contrast, MiniLM tends to emphasize semantic or functional relatedness: for \textit{clipboards}, it retrieves \textit{paper clip} and \textit{post-it notes}, and for \textit{fork}, it retrieves \textit{spoon} and \textit{knives}. 
This comparison highlights that different text encoders induce different neighborhood structures, yet both provide semantically organized anchor spaces. 
% This helps explain why our framework is not tied to a specific text encoder, while also showing that the structure of the anchor space can influence which relations are emphasized.

% \section{Open-Set Generalization}
% \label{sec:app:os}
% \begin{table}[h]
% \footnotesize
% \centering
% \caption{Comparison on Open-Set DG.}
% \label{exp:open_set}
% \begin{tabular}{l|cc}
% \toprule
%         & Known classes & Unseen classes \\
% \midrule
% CLIP    & 86.1 & 77.6 \\
% CLIPood & 89.4 & 78.2 \\
% \rowcolor[rgb]{0.949,0.949,0.949} Ours & \textbf{90.6} & \textbf{78.8} \\
% \bottomrule
% \end{tabular}
% \end{table}

% We further examine open-set domain generalization, which evaluates a model's ability to generalize across domains and to unseen classes beyond the trained ones—a more realistic and challenging setting.
% Following~\cite{pmlr-v202-shu23a}, we separately measure accuracy on the seen half of OfficeHome classes used during training and on the unseen half to assess how well competing methods generalize.
% As shown in \cref{exp:open_set}, our method outperforms baselines, benefiting from the regularization that aligns the embedding space to preserve semantics while filtering domain-specific noise.

\end{document}